\documentclass[10pt,twocolumn,letterpaper]{article}

\usepackage{cvpr}
\usepackage{times}
\usepackage{epsfig}
\usepackage{graphicx}
\usepackage{amsmath}
\usepackage{amssymb}
\usepackage{caption}
\usepackage{subfig}
\usepackage{placeins}
\usepackage[dvipsnames]{xcolor}
\usepackage{appendix}
\usepackage{float}

\usepackage[pagebackref=true,breaklinks=true,letterpaper=true,colorlinks,bookmarks=false]{hyperref}

\cvprfinalcopy % *** Uncomment this line for the final submission

\def\cvprPaperID{4196} % *** Enter the CVPR Paper ID here

\ifcvprfinal\fi
\begin{document}

%%%%%%%%% TITLE
\title{Multiple People Tracking Using Hierarchical Deep Tracklet Re-identification}

\author{Maryam Babaee\thanks{equal contribution} \qquad Ali Athar\footnotemark[1] \qquad Gerhard Rigoll\\
Institute for Human-Machine Communication, Technical University of Munich\\
Arcisstrasse 21, Munich, Germany\\
{\tt\small \{maryam.babaee@tum.de, ali.athar@tum.de, rigoll@tum.de\}}
}

\maketitle
\newcommand{\visfeature}[2]{\ensuremath{\psi(#1,#2)}}
\newcommand{\stfeature}[1]{\ensuremath{\sigma(#1)}}
\newcommand{\stfout}[2]{\ensuremath{\phi_f(#1,#2)}}
\newcommand{\strout}[2]{\ensuremath{\phi_r(#1,#2)}}
\newcommand{\tsim}[2]{\ensuremath{\alpha(#1,#2)}}
\newcommand{\cfeaturef}[2]{\ensuremath{\Omega_f(#1,#2)}}
\newcommand{\cfeaturer}[2]{\ensuremath{\Omega_r(#1,#2)}}

%%%%%%%%% ABSTRACT
\begin{abstract}
The task of multiple people tracking in monocular videos is challenging because of the numerous difficulties involved: occlusions, varying environments, crowded scenes, camera parameters and motion. In the tracking-by-detection paradigm, most approaches adopt person re-identification techniques based on computing the pairwise similarity between detections. However, these techniques are less effective in handling long-term occlusions. By contrast, tracklet (a sequence of detections) re-identification can improve association accuracy since tracklets offer a richer set of visual appearance and spatio-temporal cues. In this paper, we propose a tracking framework that employs a hierarchical clustering mechanism for merging tracklets. To this end, tracklet re-identification is performed by utilizing a novel multi-stage deep network that can jointly reason about the visual appearance and spatio-temporal properties of a pair of tracklets, thereby providing a robust measure of affinity. Experimental results on the challenging MOT16 and MOT17 benchmarks show that our method significantly outperforms state-of-the-arts.
\end{abstract}
%%%%%%%%% BODY TEXT 
\section{Introduction}

Multi-object tracking (MOT) is a key problem in computer vision with many applications such as video surveillance, activity analysis, and abnormality detection~\cite{alahi2014social,aggarwal2011human,aviv1997abnormality}. It is challenging in unconstrained environments due to influencing factors such as illumination variance, camera motion, target interactions, and more importantly, lengthy occlusions.

Most existing multi-object tracking methods fall into the tracking-by-detection category, where the goal is to link detections in the video belonging to the same target. Recent tracking methods adopt person re-identification techniques based on pairwise similarity of detections~\cite{tang2017multiple, leal2016learning} for this data association. However, this can lead to wrong associations, especially if there are lengthy occlusions. By contrast, considering a group of detections before and after an occlusion as a tracklet can improve the re-identification accuracy.
%To this end, we propose a new method for comparing two tracklets (tracklet re-identification), 
%A key challenge in pedestrian re-identification is the ability to either predict or evaluate the motion trajectory of multiple persons simultaneously. This is a non-trivial task as there are numerous associated problems. Pedestrians frequently become hidden from view for several frames due to occlusions. Additionally, since we can only observe the projection of the motion on a 2D image plane, the motion model becomes heavily dependent on the camera's intrinsic and extrinsic parameters such as its viewing angle and motion. 
\begin{figure}[t!]
\centering
\includegraphics[width=\linewidth]{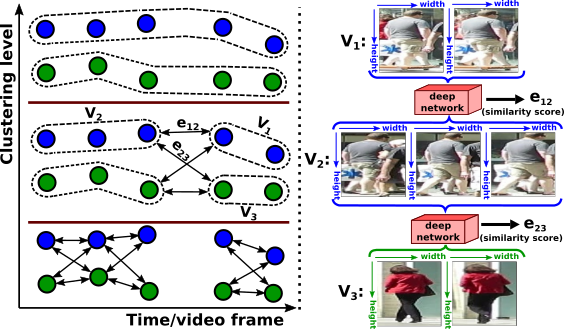}
\caption{Overview of the proposed people tracking framework that hierarchically clusters tracklets and employs a deep network to evaluate tracklet similarity. Circles denote vertices in the graph and their colors reflect their person IDs.}
\label{fig:overview}
\vspace{-5pt}
\end{figure} 

Moreover, we argue that these association errors can be further reduced if pedestrian tracking is formulated as a hierarchical clustering problem that iteratively merges detections into longer tracklets. This way, the association complexity increases gradually as opposed to a one-step approach that directly aims to obtain the final solution.

When evaluating possible associations between detections in crowded scenes where multiple pedestrians are closely located and/or overlapping in the image, it is essential to jointly reason about both their visual appearance and spatio-temporal properties. Recently, several works~\cite{quadcnn2017, headbodyfusion2018, tang2017multiple} use neural networks to process visual appearance, and separately compute hand-crafted features to incorporate spatio-temporal information from the bounding boxes. Logistic regression or some learning technique is then used to assign weights to these features in order to compute an overall similarity metric. Even though this mitigates the need to empirically set weights through trial-and-error, we argue that hand-crafted features nonetheless do not generalize well since they make certain assumptions about the underlying motion model, and as in~\cite{quadcnn2017}, some of these features may have to be computed separately for each video sequence to account for the difference in camera parameters. Additionally, such approaches lack the ability to jointly reason about spatio-temporal and visual cues in a strong manner since the features are computed separately and combined only at the final step.

In this paper, we propose a multiple people tracking framework (illustrated in Fig. \ref{fig:overview}) that hierarchically merges tracklets to overcome occlusions and minimize association errors. Our main contributions are: (1) A novel end-to-end deep network for assessing tracklet similarity that can jointly reason about visual and spatio-temporal cues in a generalized manner without requiring hand-crafted features and/or tunable parameters; (2) an extension of Kernighan-Lin with Joins algorithm~\cite{lmp2015} that enables the tracklet clustering problem to be formulated as a constrained minimum-cost multicut graph problem, and; (3) a new state-of-the-art in the MOT Challenge~\cite{leal2015motchallenge}.

%\begin{itemize}
%\item A novel end-to-end deep network for assessing tracklet similarity that can jointly reason about visual and spatio-temporal cues in a generalized manner without requiring hand-crafted features and/or tunable parameters.
%\item An extension of Kernighan-Lin with Joins algorithm~\cite{lmp2015} that enables the tracklet clustering problem to be directly formulated a constrained minimum-cost multicut graph problem.
%\item A new state-of-the-art in the MOT Challenge~\cite{leal2015motchallenge}.
%\end{itemize}

%In this paper, we propose a tracking framework that hierarchically merges tracklets to overcome occlusions and minimize association errors. To this end, similarity metrics between tracklets are computed using a novel deep learning model which is able to jointly reason about visual and spatio-temporal features in a strong and generalized manner. Finally, we extend the Kernighan-Lin with Joins algorithm~\cite{tang2017multiple} in order to effectively solve the tracklet clustering problem at each iteration.

%The main contributions of this paper are summarized as follows:
%\begin{itemize} 
%\item one \dots{We propose a novel deep neural network for tracklet matching by reasoning jointly on visual and spatio-temporal cues}
%\item two \dots{The extended Kernighan-Lin with Joins by adding constraints, adaptated to this application }
%\item three \dots{}
%\end{itemize}
%------------------------------------------------------------------------
\section{Related Work}

Most multi-object tracking approaches are based on the tracking-by-detection paradigm~\cite{levinkov2017joint,hofmann2013hypergraphs,geiger20143d}, where tracking is formulated as a data association problem between the detections extracted from a video using object detectors.

Data association can be performed either on individual detections~\cite{levinkov2017joint,babaee2017combined}, or a set of confident and short tracklets~\cite{wen2016exploiting,lan2018interacting} which are generated by first performing a low level data association to group detections. A well-known representation of the tracking-by-detection paradigm is to present each detection as a node in a graph, with edges representing the likelihood that connected detections belong to the same person. This data association problem can be solved using Conditional Random Field inference~\cite{yang2012online}, network flow optimization~\cite{zhang2008global, babaee2016pixel}, maximum multi-clique~\cite{dehghan2015gmmcp}, greedy algorithms~\cite{pirsiavash2011globally}, or subgraph decomposition~\cite{tang2015subgraph}. 

By learning discriminative feature representations, deep learning has enhanced many computer vision applications such as image classification~\cite{krizhevsky2012imagenet}, video background subtraction~\cite{babaee2018deep}, and pedestrian detection~\cite{ouyang2013joint}. In the context of tracking, Convolutional Neural Networks (CNN) have been utilized to learn feature representations of targets instead of using heuristic and hand-crafted features~\cite{wang2015transferring,li2016deeptrack,wojke2017simple}.
%In~\cite{wang2015transferring}, the authors use target features obtained from an offline, pre-trained CNN to perform data association in an online manner, whereas in ~\cite{li2016deeptrack}, a pre-trained CNN is tuned continuously during tracking to adapt to the appearance of tracked targets in the observed frames.
CNNs have also been utilized for modeling the similarity between a pair of detections~\cite{leal2016learning, tang2017multiple}.
%In~\cite{leal2016learning}, a Siamese CNN is proposed to learn the spatio-temporal affinity between two detections using their respective RGB images and the associated optical flows. The learned features are then combined with other contextual features using gradient boosting.
~\cite{son2017multi} models the appearance with temporal coherency by designing a quadruplet CNN. Adopting a different network structure, Milan \textit{et al.}~\cite{milan2016online} propose an end-to-end Recurrent Neural Network (RNN) for the data association problem in online multi-target tracking. They use RNNs for target state prediction, and to determine a track's birth/death in each frame.

Among other online multi-target tracking approaches which are based on tracklet-detection matching,~\cite{zhou2018online} exploits structural invariance constraint and develops a probability frame that is able to jointly reason about both appearance and structure cues for an object-detection pair. In~\cite{zhu2018online}, the authors propose an online tracking method using dual matching attention networks with both spatial and temporal attention mechanisms. 
%The spatial attention module generates dual attention maps which enable the network to focus on the matching patterns of the input image pair, while the temporal attention module adaptively allocates different levels of attention to different samples in the tracklet to suppress noisy observations. 
In~\cite{fang2018recurrent}, a temporal generative modeling framework is proposed that uses a recurrent autoregressive network to characterize the appearance and motion dynamics of multiple objects over time. In~\cite{long2018tracking}, a novel scoring function based on a fully convolutional network is presented to perform optimal selection from a large number of candidates in real-time.~\cite{chu2017online} utilizes the merits of single object trackers using shared CNN features and Region of Interest (ROI) pooling. In addition, a spatial-temporal attention mechanism was adopted to alleviate the problem of drift caused by frequent occlusions.

Recently, Ma \textit{et al.}~\cite{ma2018trajectory} presented a framework that employs a three step process in which tracklets are first created, then cleaved, and then reconnected using a combination of Siamese-trained CNNs, Bi-Gated Recurrent Unit (GRU) and LSTM cells. By contrast, our approach utilizes a hierarchical clustering mechanism with a single multi-stage network to compute tracklet similarity, thereby minimizing false associations in the first step and mitigating the need for tracklet cleaving and reconnection. In~\cite{sadeghian2017tracking}, a multi-stage network was proposed to model the appearance, motion and interaction of targets. Their network design is similar to ours, but with the key difference that our model computes the similarity between two tracklets, rather than between a tracklet and a single detection.
% ~\cite{Bewley2016} proposes a simple and real time online tracking method by employing a Kalman filter as a linear motion predictor, and the Hungarian algorithm for data association.~\cite{wojke2017simple} extends the work of~\cite{Bewley2016} by applying a deep association metric based on visual appearance.
%In general, online tracking approaches have more practical applications, but offer lower overall performance due to the inability to observe subsequent video frames at each time step.
%Different from other deep-learning based tracking frameworks, we propose a novel deep network that reasons \emph{jointly} on spatio-temporal and visual appearance cues of a pair of %\emph{tracklets}. 
%------------------------------------------------------------------------
\section{Approach}
\begin{figure*}
\centering
\includegraphics[width=\linewidth]{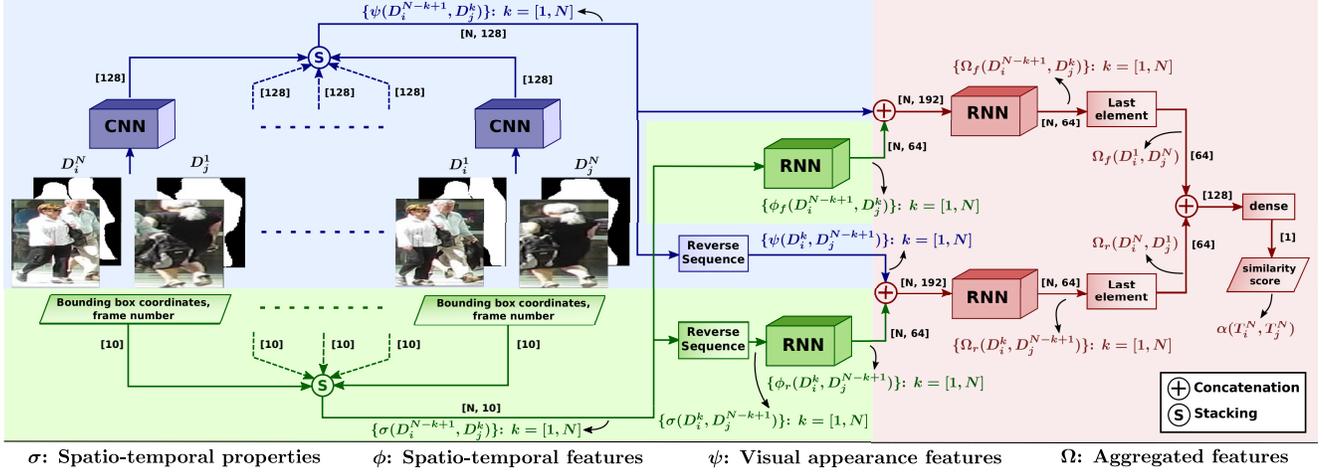}
\caption{Block diagram of the end-to-end network. The region highlighted in blue is relevant to visual appearance features, the region in green to spatio-temporal features, and the region in red to the combined representation of both. All CNN blocks share the same parameters. The feature dimensions are written in square brackets.}
\label{fig:network_block_diagram}
\end{figure*}

The proposed framework hierarchically merges tracklets to reduce association errors. It utilizes a deep network for tracklet re-identification that computes pairwise similarity scores between tracklets by jointly learning visual and spatio-temporal features. This network consists of a CNN that learns pairwise detection visual appearance, and two bidirectional RNNs that learn spatio-temporal features, and aggregate visual and spatio-temporal features, respectively. Hierarchical clustering is formulated as a series of constrained minimum cost multicut graph problems with vertices representing tracklets, and edges representing tracklet similarities as computed by the network. 
%The proposed framework primarily consists of three components: (1) a CNN that compares the visual appearance of a pair of person detection images, (2) an RNN which computes the similarity between a pair of tracklets using the visual features and bounding box coordinates of the detections that make up the tracklets, and (3) a clustering algorithm that formulates tracklet clustering as a minimum cost multicut problem on an undirected graph, and iteratively merges tracklets until convergence. These components are discussed in detail in the following subsections:

\subsection{Deep Tracklet Re-identification}
\label{sec:deep_tracklet_association}

Before elaborating the network architecture, let us define the following nomenclature: a tracklet $T_i^N$, uniquely identified by $i$, is defined as a collection of $N$ detections $\{D^1_i, D^2_i,... , D^{N-1}_i, D^N_i\}$ such that $N\in [1, N_\text{max}]$, subject to the constraint that no more than one detection is allowed in any given image frame of the video sequence. Let $F(D_a)$ denote the frame number in which detection $D_a$ lies. Also assume that the detections of $T_i^N$ are sorted in ascending order of frame number, i.e., $F(D_i^1) < F(D_i^2) < ... < F(D_i^N)$. Geometrically, a detection $D_a$ is a rectangular bounding box in the image plane that is described by the tuple $\sigma(D_a)=[X_a,Y_a,W_a,H_a,F(D_a)] \in \mathbb{R}^5$, where $X_a$ and $Y_a$ are the top-most and left-most pixel coordinates, respectively, and $W_a$ and $H_a$ are the width and height, respectively. Before any $\stfeature{.}$ is input to the network, the bounding box dimensions are normalized by the image dimensions, and offset by the coordinates of the first detection in a given tracklet pair. Similarly, the frame number of the first detection in a given tracklet pair is considered 0, and all subsequent frame numbers are normalized by the frame rate of the video.

Let us further define $\tsim{T_i^L}{T_j^M} \in [0,1]$ as the probability of tracklets $T_i^L$ and $T_j^M$ belonging to the same person. Here, we impose the constraint that $\tsim{T_i^L}{T_j^M}$ can only be computed when tracklet $T_i^L$ precedes $T_j^M$ in the video sequence with no overlapping frames, i.e. $F(D_i^1) \leq F(D_i^L) < F(D_j^1) \leq F(D_j^M)$. Since the framework processes detections pairwise, the number of detections in both tracklets is reduced to $N=\min(L,M, N_\text{max})$ by removing the first $L-N$ detections from $T_i^L$ and the last $M-N$ detections from $T_j^M$. Let us refer to these pruned versions of $T_i^L$ and $T_j^M$ as $T_i^N$ and $T_j^N$, respectively.

\subsubsection{Visual Appearance Feature Learning}
\label{sec:cnn_description}
To learn visual appearance features, we employ a CNN based on the ResNet-50 architecture~\cite{he2016deep} which compares a pair of detections and outputs the probability of those detections belonging to the same person. The input to this network is a pair of RGB detection images, along with a binary body mask for both detections that is active at pixel locations occupied by persons. The motivation behind incorporating the body mask is to focus the CNN's attention on the relevant part of the image so that it becomes more sensitive to changes in the person's appearance and learns to ignore background changes. These masks are generated using a pre-trained Mask-RCNN~\cite{he2017mask}. The RGB images and body masks are resized to 128x128. Since this dimension size is roughly half that used in~\cite{he2016deep}, the first convolutional filter of our CNN is 5x5 instead of 7x7, and a stride of 1 is used when applying this filter instead of 2. 

The input tensor dimensions are thus 128x128x8 (2x3 RGB image channels and 2x1 binary body masks). The output from the convolutional layers is flattened and input to a dense layer which reduces the feature size to 128. Given the input detection pair $(D_a, D_b)$, let us refer to this feature vector as $\visfeature{D_a}{D_b} \in \mathbb{R}^{128}$. $\visfeature{D_a}{D_b}$ is then input to a  classification layer that contains a single neuron with sigmoid activation that outputs the probability of detections $D_a$ and $D_b$ belonging to the same person.
%Qualitatively, the CNN performance is mainly effected by two factors: (1) frame gap: if the detections are only separated by a small frame gap, then the pose and appearance of the person usually does not change much, and so the CNN is able to associate them accurately; and (2) lighting conditions: large exposure differences are challenging for the CNN to identify. This occurs frequently in daytime outdoor videos with bright sunlight.

\begin{figure}[t]
\centering
\includegraphics[width=\linewidth]{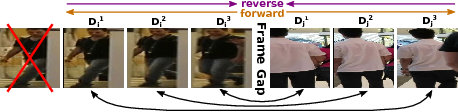}
\caption{Illustration of the pairwise detections and their ordering for two example tracklets $T_i^L$ (left) and $T_j^M$ (right) with $L=4, M=3$ separated by a frame gap. The first detection of $T_i^L$ is pruned since $N=\min(4,3)=3$ (ignoring $N_\text{max}$).}
\label{fig:ordering}
\vspace{-8pt}
\end{figure}

To compute $\tsim{T_i^L}{T_j^M}$, the first step is to apply the CNN to compute the pairwise similarity features for the following sequence of detection image pairs in $T_i^N$ and $T_j^N$: $\{(D_i^{N-k+1},D_j^k)\}$ for $k \in [1,N]$. 
%$(D_i^N, D_j^1), (D_i^{N-1}, D_j^2), ..., (D_i^2, D_j^{N-1}), (D_i^1, D_j^N)$. 
This results in a sequence of $N$ feature vectors $\{\visfeature{D_i^{N-k+1}}{D_j^k}\}$ for $k \in [1,N]$. An illustration of the detection pairs in a pair of tracklets is given in Fig. \ref{fig:ordering} using black, curved arrows.
%$\visfeature{D_i^N}{D_j^1}, \visfeature{D_i^{N-1}}{D_j^2}, ..., \visfeature{D_i^2}{D_j^{N-1}}, \visfeature{D_i^1}{D_j^N} \allowbreak \in \mathbb{R}^{N %\times 128}$.

\subsubsection{Spatio-temporal Feature Learning}

Separately, the sequence of spatio-temporal properties of the detection pairs belonging to tracklets $T_i^N$ and $T_j^N$ is input to a bidirectional RNN. Formally speaking, the sequence $\{\stfeature{D_i^{N-k+1}} \oplus \stfeature{D_j^k}\}$ for $k \in [1,N]$ (where $\oplus$ denotes concatenation of two vectors), and its reversed version $\{\stfeature{D_i^k} \oplus \stfeature{D_j^{N-k+1}}\}$ for $k \in [1,N]$, are input to two separate series of GRU cells of size 64, resulting in the two output sequences $\{\stfout{D_i^{N-k+1}}{D_j^k}\}$ and $\{\strout{D_i^k}{D_j^{N-k+1}}\}$ for $k \in [1,N]$, respectively. Intuitively, these sequences encode the spatio-temporal features of the pairwise combinations of detection bounding boxes in tracklets $T_i^N$ and $T_j^N$. Note that while the visual appearance features are independent of the sequence in which the pairs of detections occur, spatio-temporal features are not. Fig. \ref{fig:ordering} illustrates the direction of the forward and backward sequences.

\subsubsection{Feature Aggregation}

%$\visfeature{D_i^N}{D_j^1} \oplus \stfout{D_i^N}{D_j^1}, ..., \visfeature{D_i^1}{D_j^N} \oplus \stfout{D_i^1}{D_j^N} \in \mathbb{R}^{N \times 192}$
The visual and spatio-temporal features of the tracklets are then concatenated, and input to another bidirectional RNN. Formally speaking, the sequences $\{\visfeature{D_i^{N-k+1}}{D_j^k} \oplus \stfout{D_i^{N-k+1}}{D_j^k}\}$ and $\{\visfeature{D_i^k}{D_j^{N-k+1}} \oplus \strout{D_i^k}{D_j^{N-k+1}}\}$ for $k \in [1,N]$ are input to two series of GRU cells of size 64, resulting in the two output sequences $\{\cfeaturef{D_i^{N-k+1}}{D_j^k}\}$ and $\{\cfeaturer{D_i^k}{D_j^{N-k+1}}\}$ for $k \in [1,N]$, respectively. Intuitively, these features offer a combined representation of the visual and spatio-temporal features of the detections in the tracklets. Since we are interested in a single similarity score that considers both sequences in their entirety, we retain only the last two elements of the output, i.e.,  $\cfeaturef{D_i^1}{D_j^N}$ and $\cfeaturer{D_i^N}{D_j^1}$. Even though it is not reflected in the notation used, these two features actually incorporate information from the entire sequence of detections, because the input to an RNN cell consists of the input at the current time-step, as well as the output from the cell at the previous time-step. Finally, $\cfeaturef{D_i^1}{D_j^N}$ and $\cfeaturer{D_i^N}{D_j^1}$ are concatenated and input to a single neuron with sigmoid activation that outputs the final similarity score $\tsim{T_i^L}{T_j^M}$. A block diagram of the complete network is provided in Fig. \ref{fig:network_block_diagram}.

\subsection{Hierarchical Clustering}
\label{sec:hierarchical_clustering}

The task of clustering tracklets globally given their pairwise similarities is formulated as a minimum cost multicut graph problem (MP)~\cite{mp1993, mp1997}. Given a graph $G=(V,E)$, tracklets are modeled as vertices $V$ in the graph, and undirected edges $E$ allow pairs of tracklets to be checked for similarity and merged. Letting $c_e$ denote the cost of an edge $e$, the MP can be defined as:

\vspace{-3pt}
\begin{align}
Y^*_E = \underset{y \in Y_E}{\text{min}}&\hspace{3pt} \sum_{e \in E} c_ey_e\\
\text{s.t.} \hspace{2pt}\forall C \in \text{cycles}(G), \forall e& \in C: y_e \leq \sum_{e' \in C \char`\\ \{e\}}y_{e'} \label{eq:transivity_constraints}
\end{align}
\vspace{-3pt}

where $y_e \in \{0,1\}$ is a 01-label assigned to edge $e$. A zero indicates a 'join', i.e., the vertices connected by the edge belong to the same component, whereas a one indicates a 'cut', i.e., the vertices belong to separate components. The output of the MP, $Y^*_E$, defines a valid decomposition of the graph into one or more disjoint components, such that the sum of costs of the cut edges is minimized (note that the number of resulting components does not have to be specified in advance). Eq. (\ref{eq:transivity_constraints}) defines transitivity constraints which guarantee that the decomposition is well-defined. It follows that if edge costs denote the similarity between vertices, then the MP can be directly applied to the tracklet clustering problem.

\begin{figure}
\centering
\includegraphics[width=\linewidth]{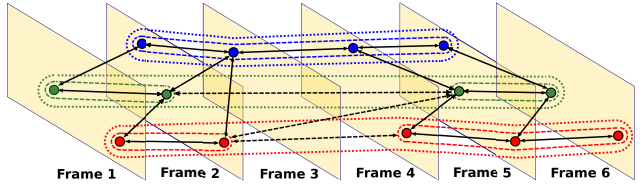}
\caption{Illustration of the hierarchical clustering process. Detections are denoted by circles, and each color corresponds to a person ID. In the first iteration, edges between adjacent detections are created (shown by solid black arrows). The resulting clusters shown by the dashed colored lines form the vertices for the second iteration, in which longer edges shown by the dashed black arrows are created. The clustering result of the second iteration is shown by the outer-most dotted lines.}
\label{fig:basic_clustering}
\end{figure}

Initially, all detections in the video sequence are assumed to be separate tracklets/vertices (both terms will be used inter-changeably from here onward). Edges are then created between vertices in adjacent frames, and their costs are computed using the similarity score obtained from the visual appearance matching network described in Sec. \ref{sec:cnn_description}. We then apply the Constrained Kernighan-Lin with Joins algorithm (described ahead) to compute a feasible decomposition of this graph; all detections that belong to the same component are subsequently merged into a single, longer tracklet. For the next iteration, edges are created between these newly merged tracklets, and the graph is again decomposed using the minimum cost multicut algorithm. Likewise, this process repeats until no more tracklets can be merged. As the tracklets become longer in subsequent iterations, we allow longer edges to be created that span over increasingly larger frame gaps in order to overcome occlusions. Moreover, for tracklets containing more than one detection, the complete network is used to compute the similarity score. This design choice will be justified in Sec. \ref{sec:ablation_study}, but the underlying idea is that once tracklets become longer, the complete network offers improved performance since the RNN is able to leverage sequential patterns in the learned features. An abstract example of the clustering process is provided in Fig. \ref{fig:basic_clustering}. Lastly, note that the tracklet similarity scores $\in [0,1]$ output by the network are mapped onto the range $[-\infty,\infty]$ by the following function to obtain the edge costs. This results in dissimilar edges having negative costs, encouraging the algorithm to cut them. 

\setlength{\abovedisplayskip}{3pt}
\setlength{\belowdisplayskip}{10pt}
\begin{align}
\label{eq:edge_weight_mapping}
f(x) = \log\left( \frac{x}{1-x} \right)
\end{align}

\subsubsection*{Constrained Kernighan-Lin with Joins}

Since the MP is NP-hard~\cite{bansal2004correlation, MPNPhard2006}, it is normally not feasible to compute a globally optimal solution. In~\cite{lmp2015}, a generalization of the MP, namely the minimum cost Lifted Multicut Problem (LMP), is proposed, and an extension of the Kernighan-Lin algorithm~\cite{kernighan1970}, called Kernighan-Lin with Joins (KLJ), is presented to solve the problem. 
%In~\cite{tang2016multi}, the task of multiple people tracking is formulated as an MP, and solved using the KLJ algorithm. 
In this work, we propose a straightforward extension of the KLJ algorithm called Constrained Kernighan-Lin with Joins (CKLJ), and employ it to solve the MP.
%Note that we do not formulate the clustering problem as an LMP, since the main reason behind using lifted edges in~\cite{tang2017multiple} was to overcome occlusions and identity switches, whereas in our framework these are already handled by the RNN.

\begin{figure}
\centering
  \centering
  \subfloat[]{\label{fig:cklj_example_a}\includegraphics[width=0.13\textwidth]{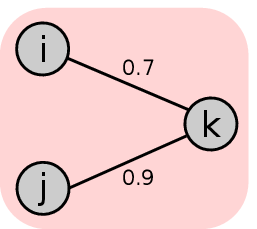}}
  \qquad
  \subfloat[]{\label{fig:cklj_example_b}\includegraphics[width=0.13\textwidth]{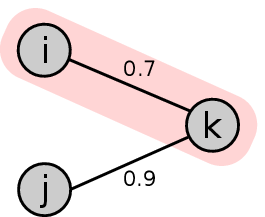}}
  \qquad
  \subfloat[]{\label{fig:cklj_example_c}\includegraphics[width=0.13\textwidth]{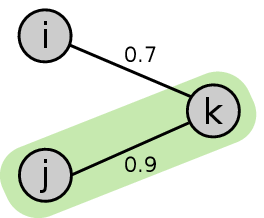}}

\caption{Illustration of the motivation behind extending the KLJ algorithm. The circles represent the three tracklets with single detections. The probability scores have been directly shown as edge weights here for ease of understanding.}
\label{fig:cklj_example}
\end{figure}

The motivation behind extending the algorithm is that when KLJ is applied to the tracklet clustering problem, it often outputs invalid results where multiple tracklets are assigned to the same component even though some of their detections lie in the same frame. As a basic example, consider three tracklets $T_i^1$, $T_j^1$ and $T_k^1$ that contain only a single detection. Suppose that $T_i^1$ and $T_j^1$ lie in the first frame of the video whereas $T_k^1$ lies in the second frame. Now, if we create edges $(T_i^1$, $T_k^1)$, and $(T_j^1, T_k^1)$, and if, for some reason (appearance similarity or spatio-temporal proximity), the similarity scores $\tsim{T_i^1}{T_k^1}$ and $\tsim{T_j^1}{T_k^1}$ are both high, then the KLJ algorithm will not cut either edge, resulting in all three detections being assigned to the same component, as illustrated in Fig. \ref{fig:cklj_example_a}. Note that this happened even though there is no direct edge between $T_i^1$ and $T_j^1$. To overcome this, CKLJ accepts a set of constraint pairs as input, where each pair $(a,b)$ defines a constraint that tracklets $T_a$ and $T_b$ cannot be assigned to the same component. Since the KLJ algorithm reduces the total cost by greedily merging/splitting components and swapping vertices between them, such constraints can be easily incorporated by imposing a conditional check prior to executing these transformations. For the current example, we would thus provide $(i,j)$ as a constraint to the algorithm.

Applying such constraints, however, gives rise to a new problem. Referring to the same example again, suppose that $\tsim{T_i^1}{T_k^1} = 0.7$ and $\tsim{T_j^1}{T_k^1} = 0.9$. Naturally, we would want the first edge to be cut, and the second one to be retained, but with the existing KLJ implementation, this may not happen, because when the algorithm tries to lower the total cost by merging two clusters or swapping vertices between them (in the "update\_bipartition" function in Alg. 2 in~\cite{lmp2015}), it iterates through the neighboring vertices in an undefined order. Therefore, it may happen that the algorithm encounters the edge $(T_i^1, T_k^1)$ first, joins it, and then later it is forced to cut $T_j^1$ and $T_k^1$ because joining it would violate a constraint (Fig. \ref{fig:cklj_example_b} illustrates this case). As a simple remedy to this problem, we make the CKLJ algorithm greedy by first sorting the neighboring components in descending order of the edge cost. This ensures that high similarity edges are joined first and the aforementioned scenario is avoided, as shown in Fig. \ref{fig:cklj_example_c}.

\section{Experiments}
%In this section, we provide implementation details of the framework and justify the design choices by performing various experiments.
\subsection{CNN Architecture}
To assess the performance of various CNN architectures for visual appearance matching, a test set was created using ground truth data from the MOT Benchmark 2015\footnote{\url{https://motchallenge.net/data/2D_MOT_2015}}. This set contains 7836 samples, each containing a detection pair belonging to either the same person or to different persons, and separated by different frame gaps. A CNN based on VGG-16~\cite{vgg2014} was applied, and we also trained our ResNet-50 based network on detection images without the binary body mask. As an alternative to stacking detection image pairs and inputing them to the network, we also trained a triple network~\cite{tripleNetwork2014, tripleNetwork2015} to learn discriminative visual embeddings of size 128 from the detection images individually. To this end, a network based on ResNet-50 was used to extract the embeddings, and the hinge loss function defined in~\cite{tripleNetwork2014} with a margin of 0.2 was used to train the network. An online smart mining strategy was used to create suitable training triplets based on approximate nearest neighbor search, as described in~\cite{tripleNetworkSmart2017}. To assign a binary similarity label to each sample, we computed the normalized $L_2$-distance between the feature embeddings of both detections, and selected an optimal threshold for classification across all samples (this came out to be 0.43). For the other networks, the predicted labels were obtained by thresholding the output at 0.5. The obtained results are presented in Table \ref{tab:cnn_architecture_comparison}. It is evident that the ResNet-50 based network with stacked detection images and body masks gave the highest accuracy. It also converged faster during training, and had a lower validation loss than the other networks.

\begin{table}
\centering

%% Make table font size smaller if final copy
\ifcvprfinal
\footnotesize
\else
\small
\fi

    \begin{tabular}{|l|c|c|}
    \hline
%    &\\[-9pt]
      \textbf{Network} & \textbf{Body Mask} & \textbf{Accuracy (\%)}\\[2pt]
      \hline
%      &\\[-10pt]
      Triple Network & Yes & 84.9\\
      VGG-16 & Yes & 85.4\\
      ResNet-50 & No & 86.3 \\
      ResNet-50 & Yes & 87.7\\
      \hline
    \end{tabular}
  %\caption{Classification accuracy on MOT15 training datasets for various CNNs.}
  \caption{Comparison of CNNs for detection matching accuracy.}
    \label{tab:cnn_architecture_comparison}
\vspace{-10pt}
\end{table}
\subsection{Effect of Temporal Distance on CNN Accuracy}
\label{sec:frame_gap_effect_cnn}

\begin{figure}
\centering
\includegraphics[width=0.7\linewidth]{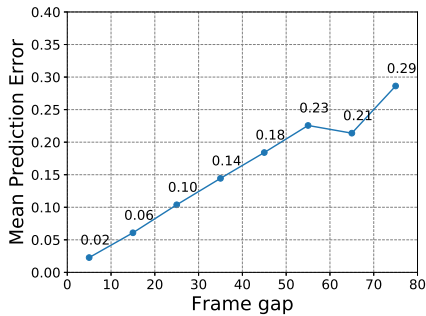}
%\caption{Effect of frame gap between detections on the classification accuracy of the CNN.}
\caption{CNN detection matching error against frame gap.}
\label{fig:cnn_frame_gap_plot}
\vspace{-10pt}
\end{figure}

One motivation behind comparing tracklets instead of detections is that the accuracy of detection matching networks deteriorates as the frame gap between them increases. To demonstrate this, we created another test set from the MOT Benchmark 2015, and plotted the mean prediction error (the average of the absolute difference between the predicted label $\in [0,1]$ output by the network and the true label $\in \{0,1\}$ for all test samples) against the frame gap in Fig. \ref{fig:cnn_frame_gap_plot}. This test set contains only same-person detection pair samples, since the visual (dis)similarity for different-person samples is largely independent of the frame gap. The results show an almost exactly linear relationship between the two parameters, which supports our claim.

\begin{figure*}[h]
\centering
  \centering
  \subfloat[]{\label{fig:rnn_accuracy_1}\includegraphics[width=0.35\textwidth]{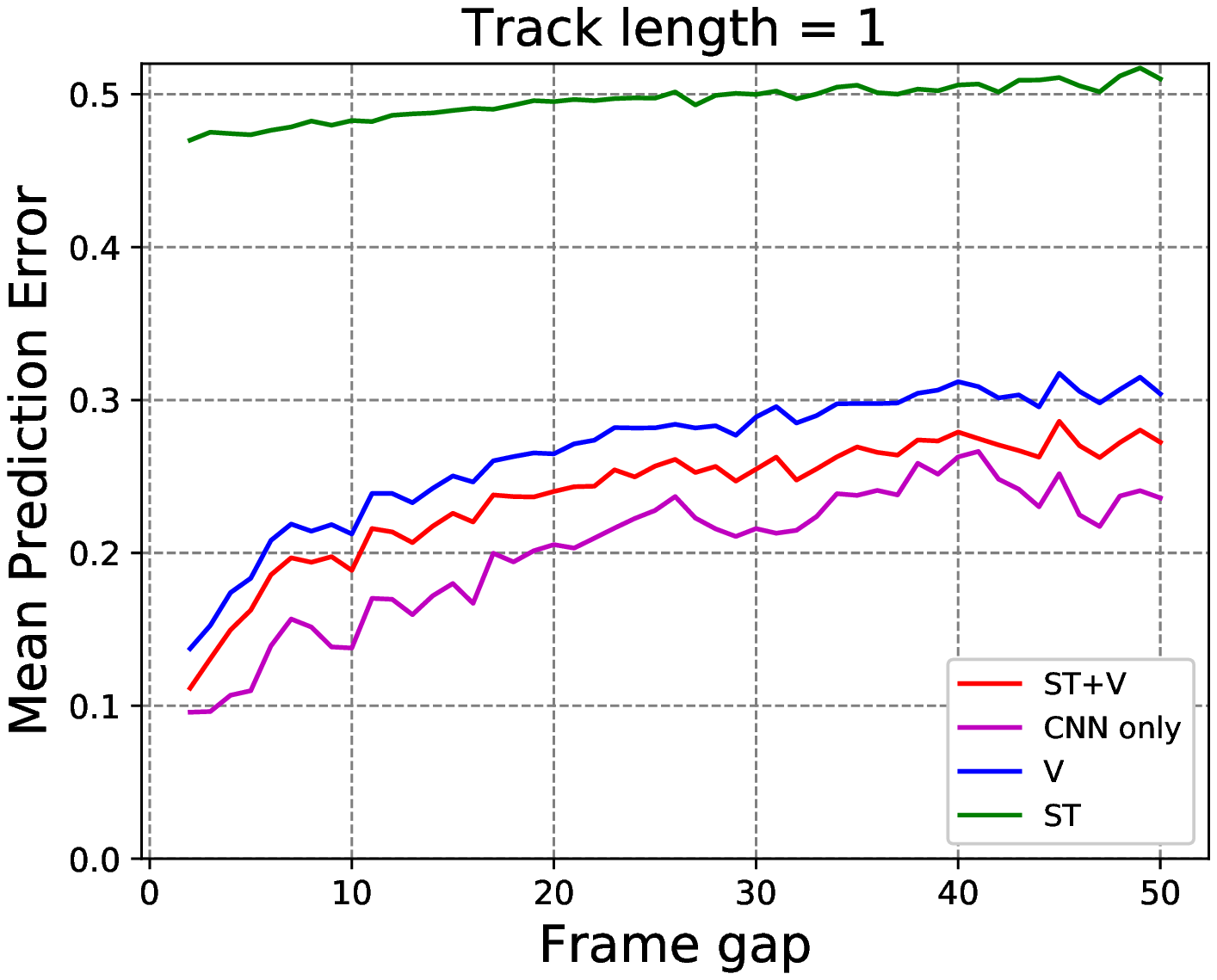}}
  \subfloat[]{\label{fig:rnn_accuracy_5}\includegraphics[width=0.35\textwidth]{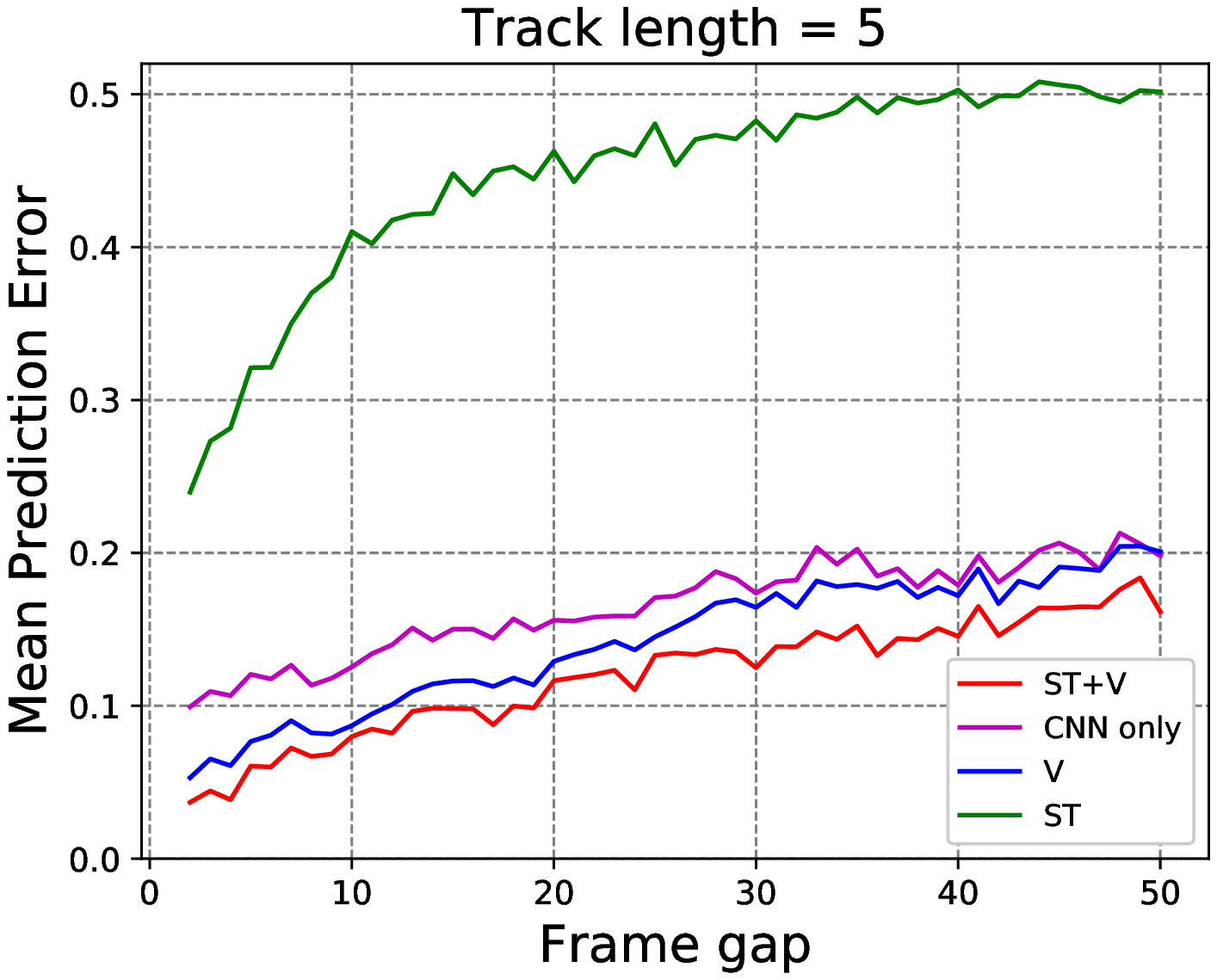}}
  \subfloat[]{\label{fig:rnn_accuracy_10}\includegraphics[width=0.35\textwidth]{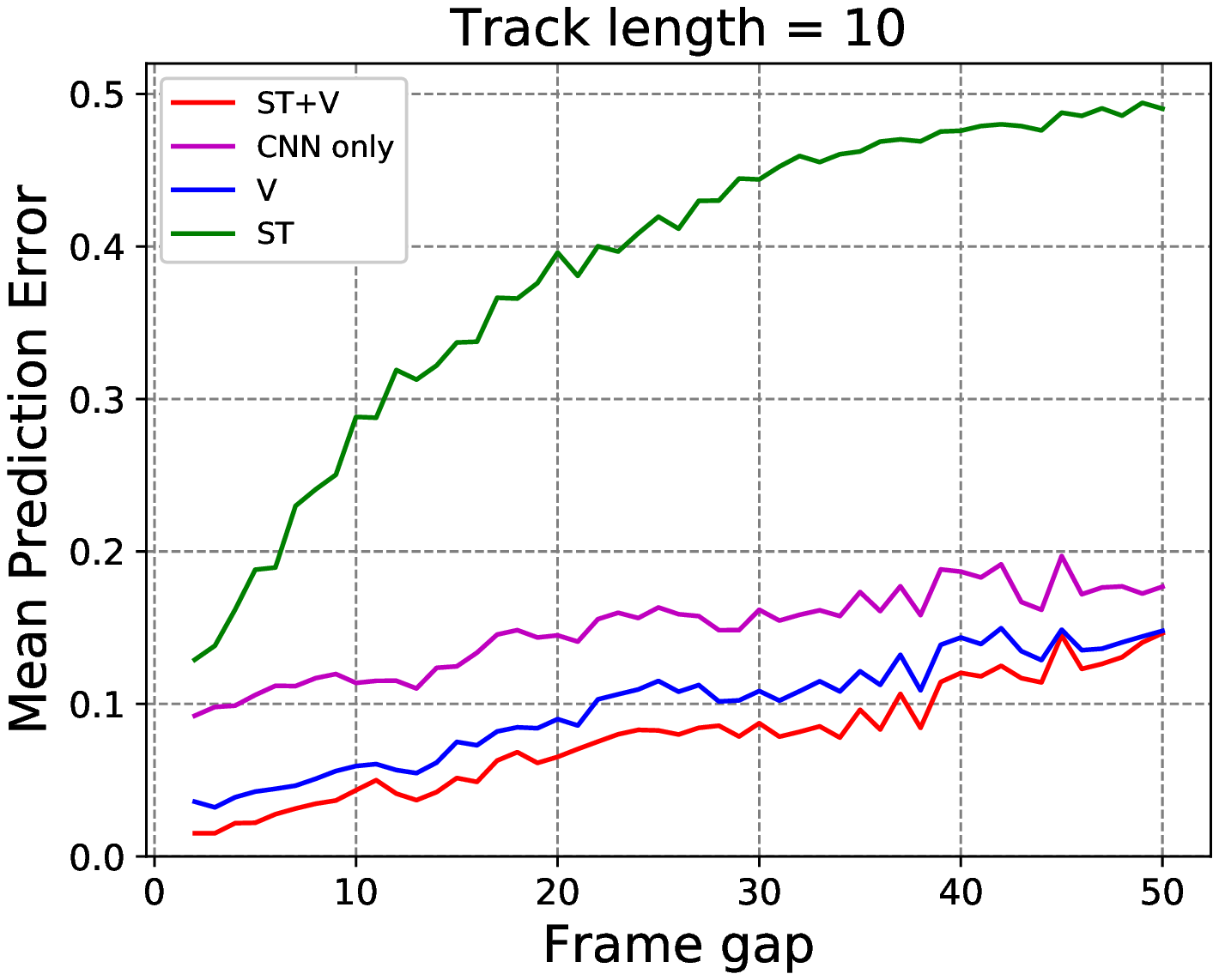}}
\caption{Mean prediction error of the RNN for various frame gaps and network inputs.}
\label{fig:rnn_accuracy}
\end{figure*}

\subsection{Ablation Study}
\label{sec:ablation_study}

To justify the use of RNNs in our tracklet matching network, we performed an ablation study in which the effect of the network's design, and the presence of visual and spatio-temporal features towards the final classification accuracy is analyzed for various track lengths and frame gaps. The different network configurations used in this study are:

\begin{itemize}
\item Spatio-temporal and visual (\textcolor{red}{ST+V}): This is the baseline configuration which uses both spatio-temporal and visual features as described in Sec. \ref{sec:deep_tracklet_association}.

\item Spatio-temporal (\textcolor{OliveGreen}{ST}): The visual features are omitted, i.e., only the spatio-temporal properties are input to the RNN.

\item Visual Sequence (\textcolor{blue}{V}): The spatio-temporal features are omitted, i.e., only the visual features output by the CNN are fed to the RNN.

\item Visual (\textcolor{magenta}{CNN only}): The RNN is omitted entirely. The tracklet similarity score is calculated by taking the mean of the similarity score output by the CNN for all detection pairs in the tracklet pair.
\end{itemize}

The RNN was trained separately for each of these configurations (except for the 'CNN only' setting), and applied to a test set created from ground truth data from the MOT Benchmark 2015. Each sample in the test set contains a pair of tracklets that may belong to the same person (positive sample) or to different persons (negative sample), and are separated by a frame gap. The length of these tracklets is either 1, 5 or 10, and there are 15000-25000 samples for each of these lengths with roughly equal positive and negative samples. In Fig. \ref{fig:rnn_accuracy}, the mean prediction error (computed in the same way as in Sec. \ref{sec:frame_gap_effect_cnn}) is plotted against frame gap for the four network configurations and three track lengths.

For tracklet length = 1, Fig. \ref{fig:rnn_accuracy_1} shows that using only the CNN similarity score results in the lowest error for all frame gaps. This is because there is almost no useful information in the spatio-temporal features if each tracklet only contains a single detection, as evident from the high error for the 'ST' configuration. In fact, the 'ST' error is approximately 0.5 regardless of the frame gap, meaning that the network's prediction is no better than a random guess. Furthermore, an RNN is only able to extract meaningful information if it has observed a longer sequence of inputs.

For tracklet length = 5 (Fig. \ref{fig:rnn_accuracy_5}), the trend changes. The spatio-temporal properties now offer useful cues to the network, as evident from the substantially lower error for the 'ST' plot for small frame gaps. Also note that the 'V' configuration outperforms the 'CNN only' configuration for all frame gaps, even though both only utilize visual appearance features. This shows that the ability of the RNN to learn sequential patterns from the input yields improved performance when a longer sequence is provided. Lastly, the baseline 'ST+V' configuration emerges as the best performer, lending credibility to our claim that an end-to-end network is able to effectively learn and aggregate different types of features. For tracklet length = 10 (Fig. \ref{fig:rnn_accuracy_10}), the same general trend continues; the error for the 'ST' configuration further reduces since spatio-temporal features become more informative, and the performance of the 'CNN only' configuration deteriorates further compared to the 'V' and 'ST+V' configurations. We conjecture that this occurs because the RNN is able to better reason about the sequential pattern of the provided features when the input sequence length is larger.

\subsection{Training}

For the CNN, a training set with 48954 detection pairs was created from ground truth data in the MOT17 training datasets. Since the ground truth detections bounding boxes are exact, we use the detection boxes output by various detectors that overlap significantly with the ground truth boxes to create training samples. This helps to make the CNN more robust to inaccurate detection boxes. To reduce over-fitting, we employ runtime image augmentation by introducing random brightness offsets and horizontal flips. Dropout~\cite{dropout2014} with keep probability of 0.8 is also applied to the final feature vector $\psi(.)$ during training. A learning rate of 0.002 with a decay factor of 0.94 after every 7000 iterations was used for optimization using Stochastic Gradient Descent (SGD) with a momentum factor of 0.9.

For the RNN, we created 22640 tracklet pairs with roughly equal numbers of positive and negative samples from the MOT17 training datasets. The samples have tracklet lengths ranging from 1 to 20, and the tracklets are separated by frame gaps ranging from 0 to 4 times the tracklet length. Positive samples are created by splitting a known ground truth track at various points. Negative samples are created in three ways: (1) a portion of another person's track is extracted such that the bounding box center coordinates of the detections of this track are closest to that of the original person's detections in the same frame. This improves the network's performance in cases where the spatial coordinate information is ambiguous. Around 50\% of negative samples are created in this manner. (2) Of the remaining, 25\% are created by dividing the image plane into four equally sized quadrants, and sampling a portion of another person's track such that the detection centers lie in the same quadrant as that of the original person. (3) The final 25\% are created similarly, but by sampling from a track whose detections lie in any other quadrant (i.e., these are easy negatives). All three sample creation techniques are detailed in App.~\ref{app:rnn_neg_sample_gen}.

When training the RNN, the weights of the CNN are frozen due to memory constraints. The RNN is trained with a learning rate of 0.002 with a decay factor of 0.95 after every 2000 iterations, and optimized using SGD with a momentum factor of 0.9. Dropout~\cite{dropout2014} with keep probability 0.5 is also applied to prevent over-fitting.

\begin{table*}[h]
\centering

%% Make table font size smaller if final copy
\ifcvprfinal
\footnotesize
\else
\small
\fi

\begin{tabular}{lcccccccccc}
\hline
Tracker & MOTA$(\%)\uparrow$  & MOTP$(\%)\uparrow$   &FAF$\downarrow$  & MT$(\%)\uparrow$  & ML$(\%)\downarrow$ &FP$\downarrow$  &FN$\downarrow$  & ID Sw.$\downarrow$   & Frag$\downarrow$ \\[1ex]
\hline 
LMP~\cite{tang2017multiple}	        & \textcolor{blue}{48.8}	 & \textcolor{blue}{79.0}	 & 1.1 &	18.2 &  	40.1 & 	6654 &	86245 &	481 &	\textcolor{blue}{595}      \\
GCRA~\cite{ma2018trajectory}        & 48.2	 & 77.5  &	0.9 &	12.9 & 	41.1 & 	5104 &	88586  & 	821 &	1117   \\ 
FWT~\cite{headbodyfusion2018}       & 47.8   &	75.5 &	1.5 &	19.1 & 	\textcolor{blue}{38.2} & 	8886 &	85487 &	852 &	1534	        \\ 
MOTDT~\cite{long2018tracking}       & 47.6	 & 74.8  & 	1.6 &	15.2 & 	38.3 & 	9253 &	\textcolor{blue}{85431} &	792 &	1858  \\  
NLLMPa~\cite{levinkov2017joint}     & 47.6   &	78.5 &	1.0 &	17.0 & 	40.4 & 	5844 &	89093 &	629 &	768  \\
AMIR~\cite{sadeghian2017tracking}   & 47.2	 & 75.8	 & \textbf{0.5} &	14.0 & 	41.6 & 	\textbf{2681} &	92856 &	774	 & 1675	   \\ 
MCjoint~\cite{keuper2016multi}      & 47.1   & 76.3 &	1.1 &	\textcolor{blue}{20.4} & 	46.9 & 	6703 &	89368 &	\textcolor{blue}{370} &	598\\ 
NOMT~\cite{choi2015near}            & 46.4   & 76.6 &	1.6 &	18.3 & 	41.4 & 	9753 &	87565 &	\textbf{359} &	\textbf{504} \\ 
JMC~\cite{tang2016multi}            & 46.3	& 75.7 &	1.1 &	15.5 & 	39.7 & 	6373 &	90914 &	657	& 1114\\ 
%DMMOT~\cite{zhu2018online}          & 46.1	& 73.8 & 1.3 &	17.4 & 	42.7 & 	7909 &	89874 &	532 &	1616	\\
%STAM16~\cite{chu2017online}         & 46.0  &	74.9 &	1.2 &	14.6 & 	43.6 & 	6895 &	91117 &	473 &	1422 \\ 
\hline
HDTR (Ours)                      & \textbf{53.6} &	\textbf{80.8} &	  \textcolor{blue}{0.8}  &	\textbf{21.2} & 	\textbf{37.0} & \textcolor{blue}{4714}	& \textbf{79353} & 618 & 833 \\
\hline		
\end{tabular}
\caption{Tracking results on the MOT16 test dataset with public detections. $\uparrow$ and $\downarrow$ represent higher is better and lower is better, respectively. The values in bold and blue represent the best and second best performances, respectively.} 
\label{tb:mot16}
%\end{adjustbox}
\end{table*}

\begin{table*}[h]
\centering

%% Make table font size smaller if final copy
\ifcvprfinal
\footnotesize
\else
\small
\fi

\begin{tabular}{lcccccccccc}
\hline
Tracker & MOTA$(\%)\uparrow$  & MOTP$(\%)\uparrow$   &FAF$\downarrow$  & MT$(\%)\uparrow$  & ML$(\%)\downarrow$ &FP$\downarrow$  &FN$\downarrow$  & ID Sw.$\downarrow$   & Frag$\downarrow$ \\[1ex]
\hline 
FWT~\cite{headbodyfusion2018}       & \textcolor{blue}{51.4}   &	77.0 &	1.4 &	21.4 & \textcolor{blue}{35.2} & 	24101 &	247921 &	2648 &	4279\\ 
jCC~\cite{keuper2018motion}            & 51.2   &  75.9 &  1.5 &   20.9 &  37.0 &  25937 & 247822 &    \textbf{1802} &  2984 \\
MOTDT17~\cite{long2018tracking}     & 50.9	 &  76.6 & 	1.4 &	17.5 & 	35.7 & 	24069 &	250768 &	2474 &	5317  \\  
MHT\_DAM~\cite{kim2015multiple} & 50.7 & \textcolor{blue}{77.5} & 1.3 & 20.8 & 36.9 & 22875 & 252889 & 2314 & \textcolor{blue}{2865} \\
EDMT17~\cite{chen2017enhancing}     & 50.0 & 77.3 & 1.8 & \textcolor{blue}{21.6} & 36.3 & 32279 & \textcolor{blue}{247297} & 2264 & 3260	 \\
HAM\_SADF17~\cite{yoon2018online}    & 48.3   & 77.2  &  \textcolor{blue}{1.2} &   17.1 &  41.7 &  \textcolor{blue}{20967} & 269038 & \textcolor{blue}{1871} & 3020 \\
DMAN~\cite{zhu2018online}          & 48.2 & 75.7 & 1.5 & 19.3 & 38.3 & 26218 & 263608 &2194 & 5378\\
PHD\_GSDL17~\cite{fu2018particle} & 48.0 & 77.2 & 1.3 & 17.1 & 35.6 & 23199 & 265954 & 3998 & 8886\\
MHT\_bLSTM~\cite{kim2018multi}    & 47.5 & \textcolor{blue}{77.5} & 1.5 & 18.2& 41.7 & 25981 & 268042 & 2069 & 3124 \\
\hline
HDTR (Ours)                       &\textbf{54.1} &	\textbf{80.2} &	  \textbf{1.0}  & \textbf{23.3} & 	\textbf{34.8} & \textbf{18002} & \textbf{238818} &	1895 &	\textbf{2693} \\
\hline		
\end{tabular}
\caption{Tracking results on the MOT17 test dataset.}
\label{tb:mot17}
\end{table*}

\subsection{Clustering Scheme}

The tracklets are iteratively merged to form longer tracklets using CKLJ, as explained in Sec. \ref{sec:hierarchical_clustering}. For the first three iterations, the maximum permitted frame gap between vertices is restricted to 1, 2 and 4, respectively. Thereafter, the frame gap is allowed to be at most four times the tracklet length. When no more tracklets can be merged, the maximum allowed frame gap restriction is further relaxed to at most six times the tracklet length. Once no more tracklets can be merged under this setting, the clustering process is said to be complete. Note that this scheme and the associated parameters have been chosen ad-hoc, with the aim of balancing fast convergence and gradual relaxation of the frame gap restriction in a manner that is applicable to both stationary and moving camera videos. We also remark that changing the parameters within a reasonable range does not effect our framework's performance significantly. Details of the edge creation method employed, and a quantitative analysis of the convergence and computational time of this clustering scheme are presented in App.~\ref{app:edge_creation},~\ref{app:graph_complexity} and~\ref{app:timing_analysis}, respectively.

\subsection{Multi-object Tracking Benchmark Results}

To assess our framework's performance, we applied it to test datasets from the MOT16 and MOT17 challenge. The MOT16 test dataset contains 7 video sequences captured in different imaging conditions with varying camera motions and camera angles. The MOT17 challenge contains the same video sequences, but offers detections from 3 different person detectors. Both challenges use the CLEAR MOT performance metrics~\cite{motclear2008} to rank tracker performance, which include Multi-Object Tracking Accuracy (MOTA) and Precision (MOTP), average False Alarms per Frame (FAF), ratio of Mostly Tracked (MT) and Mostly Lost (ML) targets, False Positives (FP), False Negatives (FN), ID switches (ID Sw.) and trajectory fragmentations (Frag.).

As evident from Tables~\ref{tb:mot16} and~\ref{tb:mot17}, our tracker outperforms all other published works on both MOT16 and MOT17 challenges in terms of overall accuracy (MOTA) by an impressive margin of 4.8\% and 2.7\%, respectively. For the other metrics, it is mostly either ranked first or second. Specifically, our approach more reliably matches tracklets across occlusions, which is evident from our high MT and low ML scores, and also from the lower false negative count. Moreover, compared to~\cite{tang2017multiple} where tracking is performed in one step graph optimization by clustering detections, our approach achieves better results by hierarchically clustering tracklets. Lastly, we recognize ID switches as an area of possible improvement. These switches occur more frequently when there is significant camera motion, which makes spatio-temporal cues less reliable, thus causing the RNN performance to deteriorate. The detailed per video sequence results and annotated videos are available online\footnote{\url{https://motchallenge.net/tracker/HDTR_16}, \url{https://motchallenge.net/tracker/HDTR_17}} and in App.~\ref{app:mot16_results}.

\section{Conclusion}
We proposed a multi-object tracking framework that hierarchically merges tracklets to effectively resolve lengthy occlusions. Tracklet clustering is formulated as a constrained minimum cost multicut problem and solved using the Constrained Kernighan Lin with Joins Algorithm. To compute similarity metrics between tracklets, a novel deep network was employed that learns and jointly reasons about spatio-temporal and visual appearance features. The framework's design choices were justified by performing various experiments, and finally, its effectiveness was demonstrated by showing its state-of-the-art performance on the MOT Challenge.

\clearpage
\FloatBarrier
{\small
\bibliographystyle{ieee}
\bibliography{egbib}
}

\newcommand{\imgext}{jpg}
\newcommand{\imgwidth}{0.29}

%\clearpage
%\FloatBarrier

\appendix
\appendixpage
%\begin{appendices}

\section{RNN Training Sample Generation}
\label{app:rnn_neg_sample_gen}

\begin{figure}[t]
\centering
\includegraphics[width=0.95\linewidth]{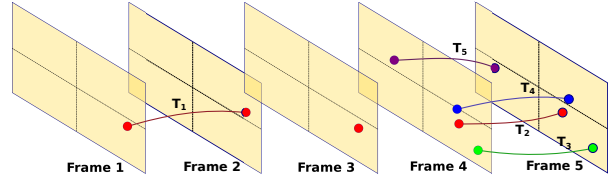}
\caption{Example of training sample creation for RNN.}
\label{fig:rnn_negative_sample_creation}
\end{figure}

An abstract example of the RNN sample generation techniques employed in the framework is given in Fig. \ref{fig:rnn_negative_sample_creation}, where samples are being created which contain two tracklets, each with two detections, and a frame gap of 1 between them. Circles denote detections, and circles of the same color belong to the same person ID. The first tracklet in the sample is $T_1$. Since a frame gap is required, the red detection in frame 3 is skipped, and tracklet $T_2$ is chosen as the second tracklet to create a positive sample. 

For negative sample creation, there are three possibilities for choosing the second tracklet: 

\begin{enumerate}
\item Selecting detections from another person's track such that these detections are spatially close to the detections in $T_2$. This is approximately achieved by searching for the detection whose bounding box center is closest to that of the red detection in frame 4, which turns out to be the blue detection. $T_4$ is therefore chosen as the second tracklet.

\item Selecting detections from another peron's track such that these detections lie in the same quadrant of the image. For this, the frame images are divided into four equally sized quadrants, as shown by the dotted lines. We then search for detections in frame 4 belonging to other perons that lie in the same quadrant as the red detection. This happens to be the detection in green, and therefore $T_3$ is chosen as the second tracklet. A random selection is made in case there are multiple candidates.

\item Selecting detections from another peron's track such that these detections lie in another quadrant of the image. The procedure for this is the same as above, except that a detection in any of the other quadrants is chosen. For this scenario, tracklet $T_5$ is a suitable candidate.
\end{enumerate}

\section{Edge Creation}
\label{app:edge_creation}

\begin{figure*}[t]
% \captionsetup[subfloat]{farskip=1pt,captionskip=1pt}
\centering
  \subfloat{\includegraphics[width=0.23\textwidth]{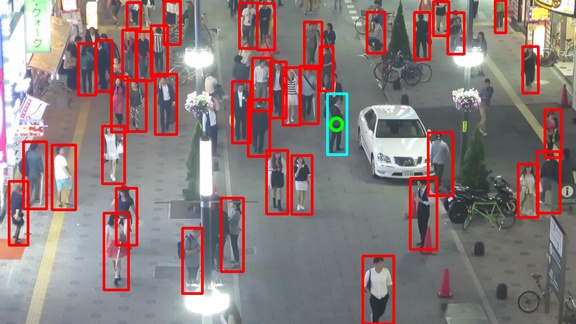}}
  \hspace{5pt}
  \subfloat{\includegraphics[width=0.23\textwidth]{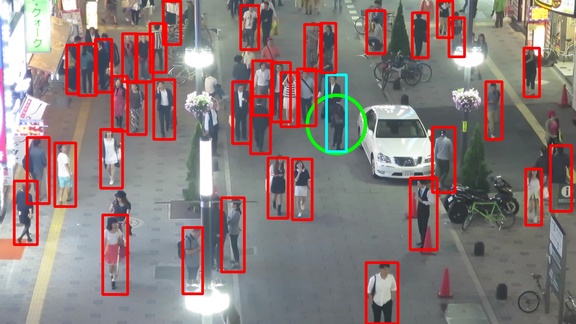}}
  \hspace{5pt}
  \subfloat{\includegraphics[width=0.23\textwidth]{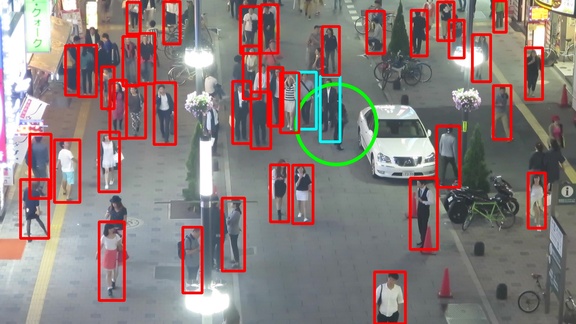}}
  \hspace{5pt}
  \subfloat{\includegraphics[width=0.23\textwidth]{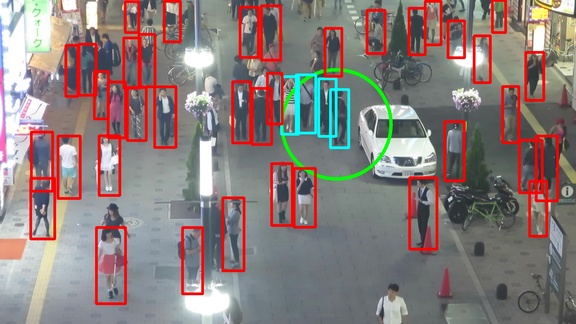}}
  \newline
  \subfloat{\hspace{-6pt}\includegraphics[width=0.23\textwidth]{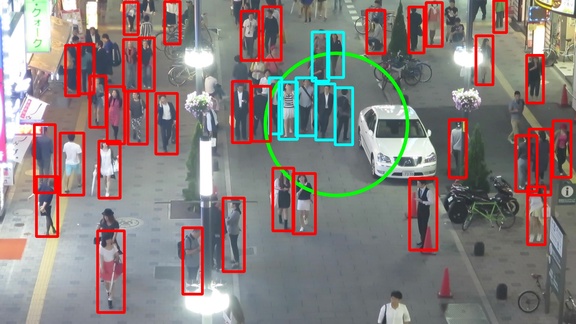}}
  \hspace{5pt}
  \subfloat{\includegraphics[width=0.23\textwidth]{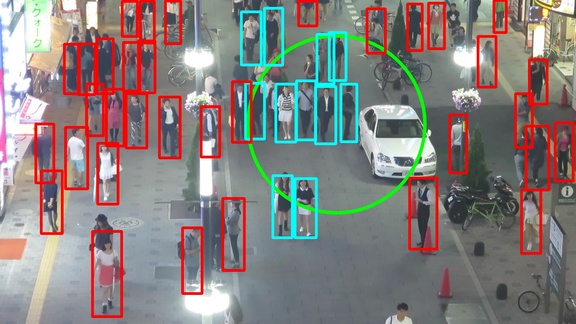}}
  \hspace{5pt}
  \subfloat{\includegraphics[width=0.23\textwidth]{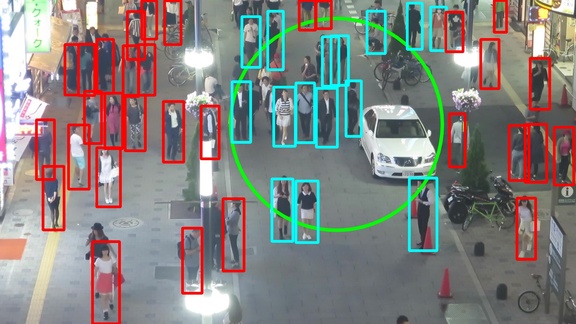}}
  \hspace{5pt}
  \subfloat{\includegraphics[width=0.23\textwidth]{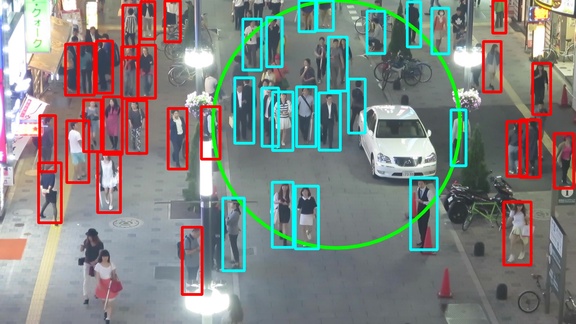}}

\caption{Visualization of the edge creation criteria. Here, we wish to find feasible edges for the detection in cyan in the upper left image. The green circle denotes the acceptable radius. The other detections in cyan are those which are accepted as edge connections, whereas those in red are not. Ordering is from left to right, and then top to bottom.}
\label{fig:edge_creation}
\end{figure*}

Since the proposed framework employs a computationally expensive neural network to compute tracklet similarities, and moreover, involves multiple iterations of graph clustering, creating edges between all combinations of tracklet pairs within the allowable frame gap range results in a very dense graph which requires a long time to evaluate. To mitigate this problem, we employ an edge creation method that avoids creating irrelevant edges. The underlying intuition is that it is very unlikely that two tracklets belong to the same person if they are spatially far apart, but temporally close. To formulate this mathematically, we compute statistics for the average, per frame bounding box movement using the provided ground-truth data in the training datasets. A separate set of statistics is computed for static and moving camera videos. Moreover, these statistics are normalized by the image dimensions, framerate, and the bounding box dimensions (larger detections are normally closer to the camera, and can therefore be expected to experience larger movements).

When applying the framework to test datasets, the pre-computed average statistics are first inflated as a safety measure, and then used to define a feasible radius around the detections of a tracklet. Edges are created only with those tracklets whose detections lie within this radius. Naturally, the radius is scaled according to the frame gap between the detections. This process is illustrated in Fig. \ref{fig:edge_creation} for the simple case where each tracklet contains a single detection. The picture sequence shows how the acceptable radius increases in size as the frame gap between the detections being considered increases. To clearly show the increasing radius, each image in the sequence occurs 10 frames after its predecessor.

It is worth pointing out that even though these statistics are heuristically computed parameters that may not generalize to all video sequences, this method of edge creation only serves to reduce computational time, and has negligible impact on the accuracy of the framework's output. This is because the edges discarded in this manner are trivial cases which the RNN can easily detect as being dissimilar. 

\section{Graph Complexity and Convergence}
\label{app:graph_complexity}

\begin{figure*}[h]
\captionsetup[subfloat]{farskip=1pt,captionskip=1pt}
\centering
  \subfloat{\includegraphics[width=0.35\textwidth]{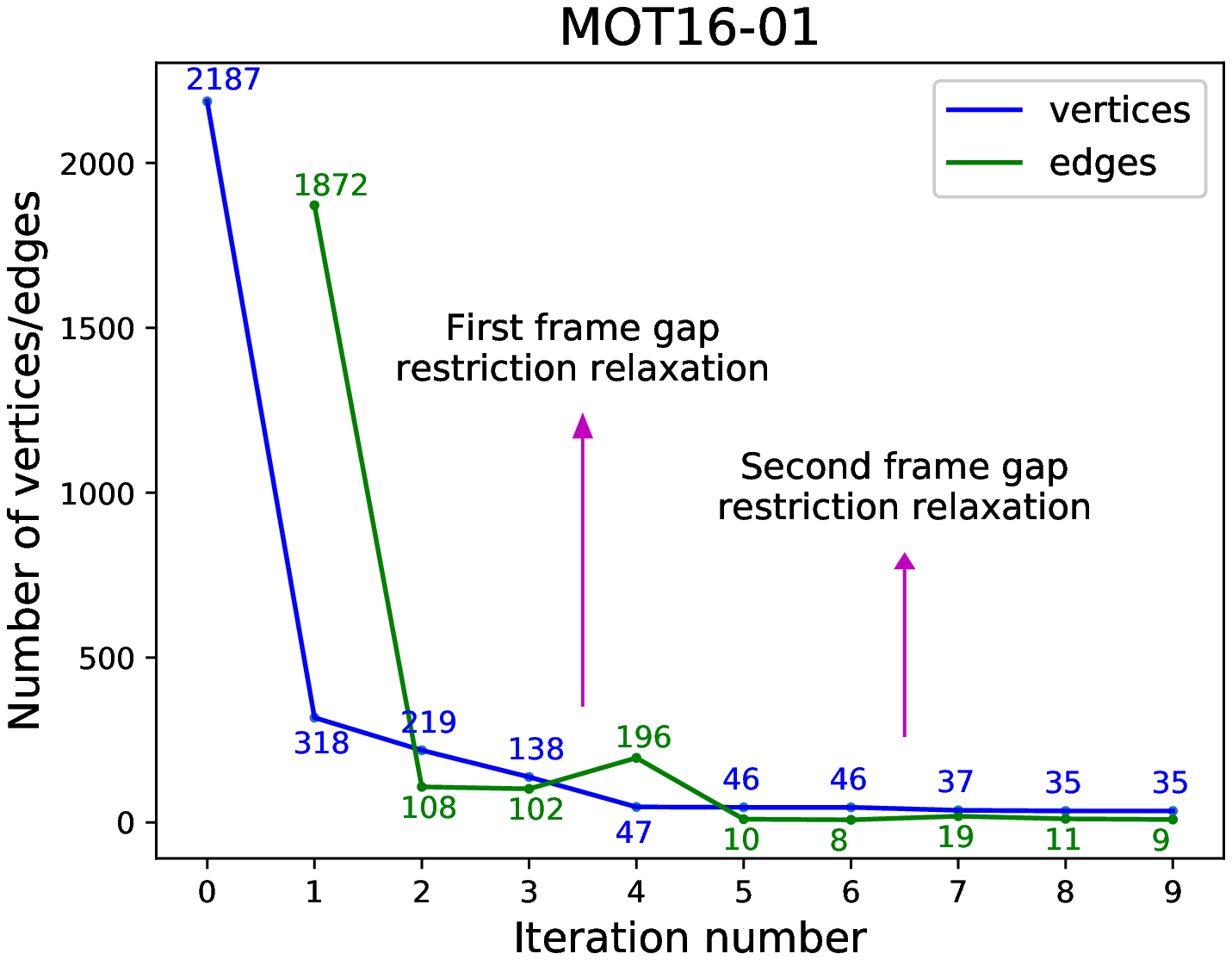}}
  \subfloat{\includegraphics[width=0.35\textwidth]{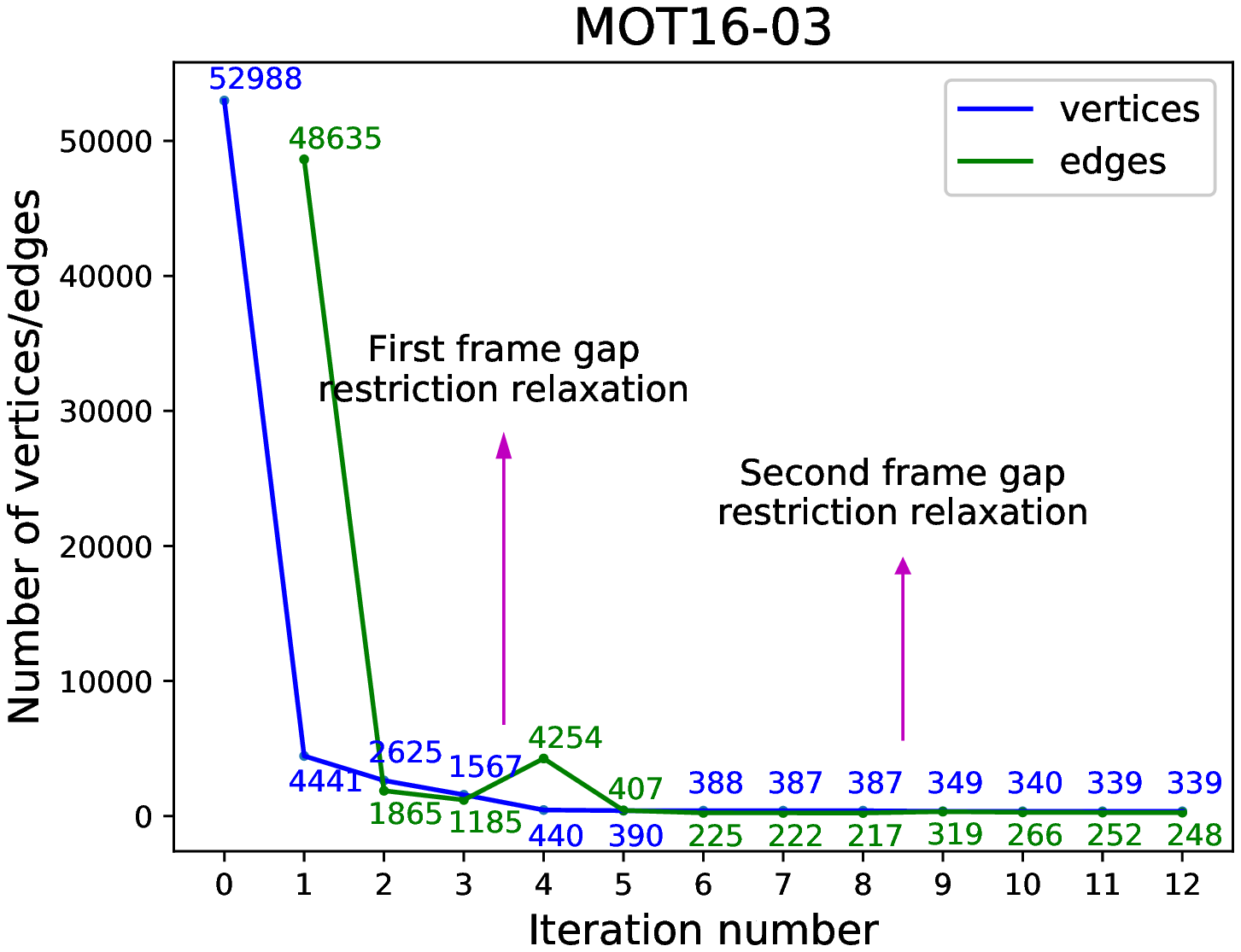}}
  \subfloat{\includegraphics[width=0.35\textwidth]{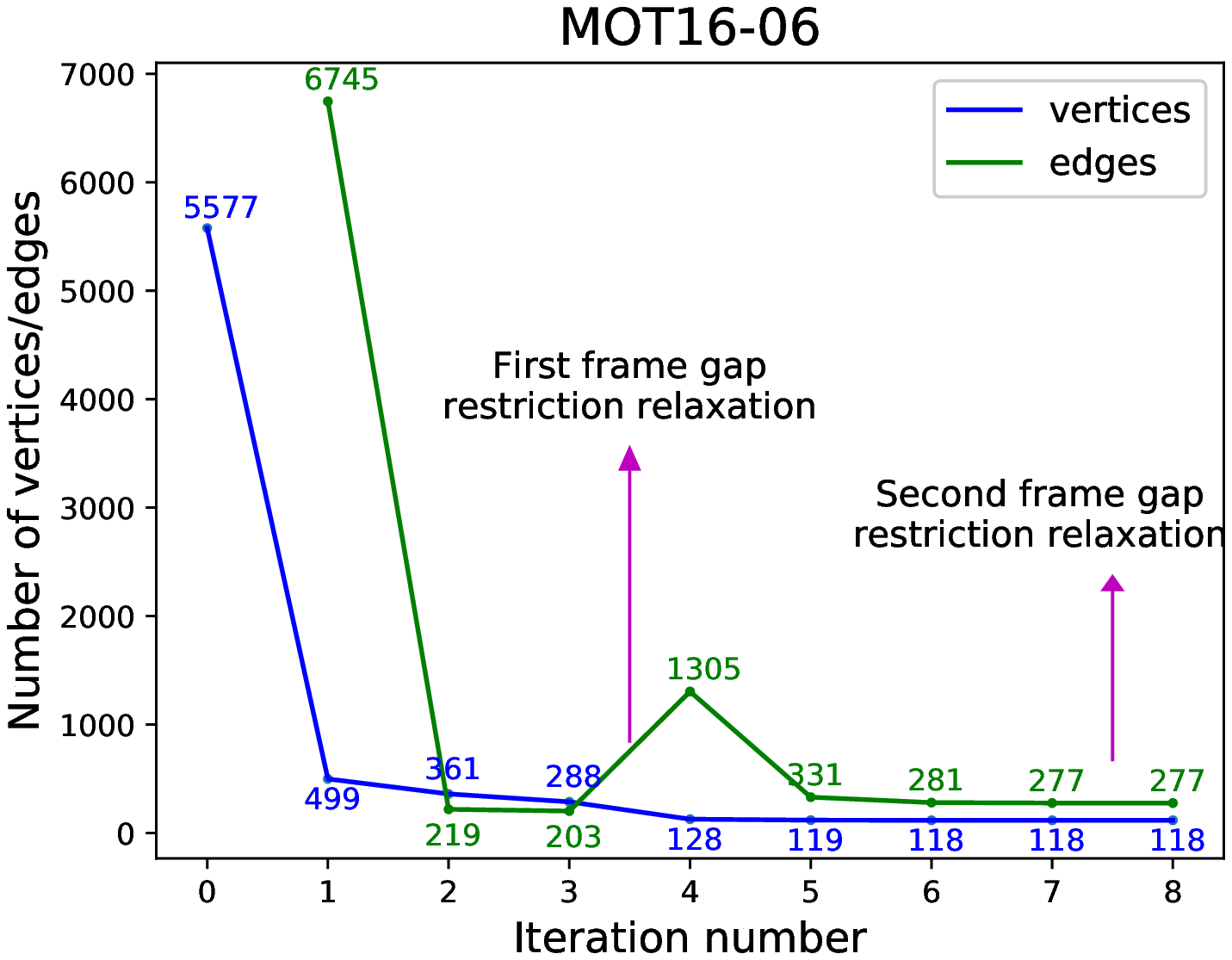}}
  \newline
  \subfloat{\includegraphics[width=0.35\textwidth]{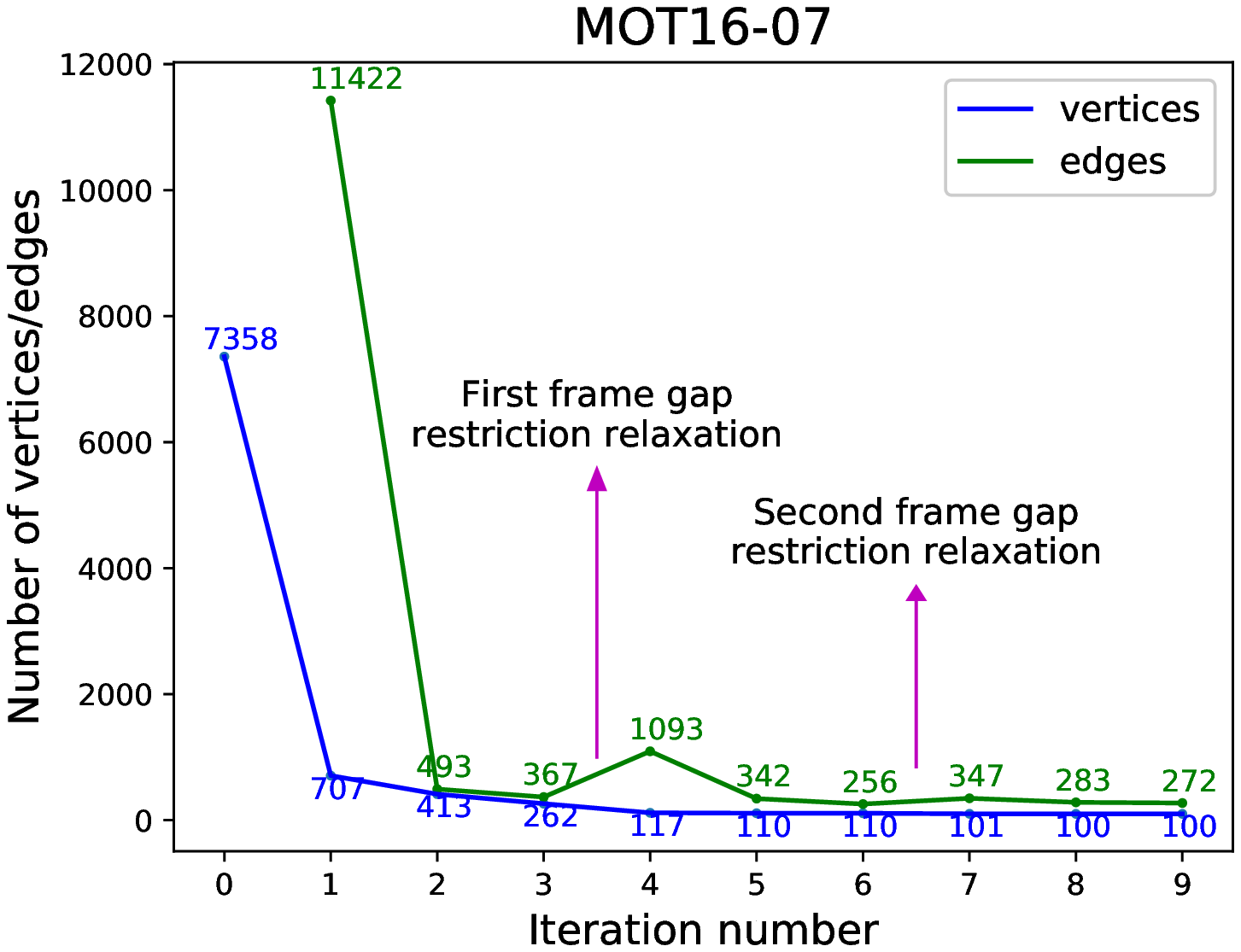}}
  \subfloat{\includegraphics[width=0.35\textwidth]{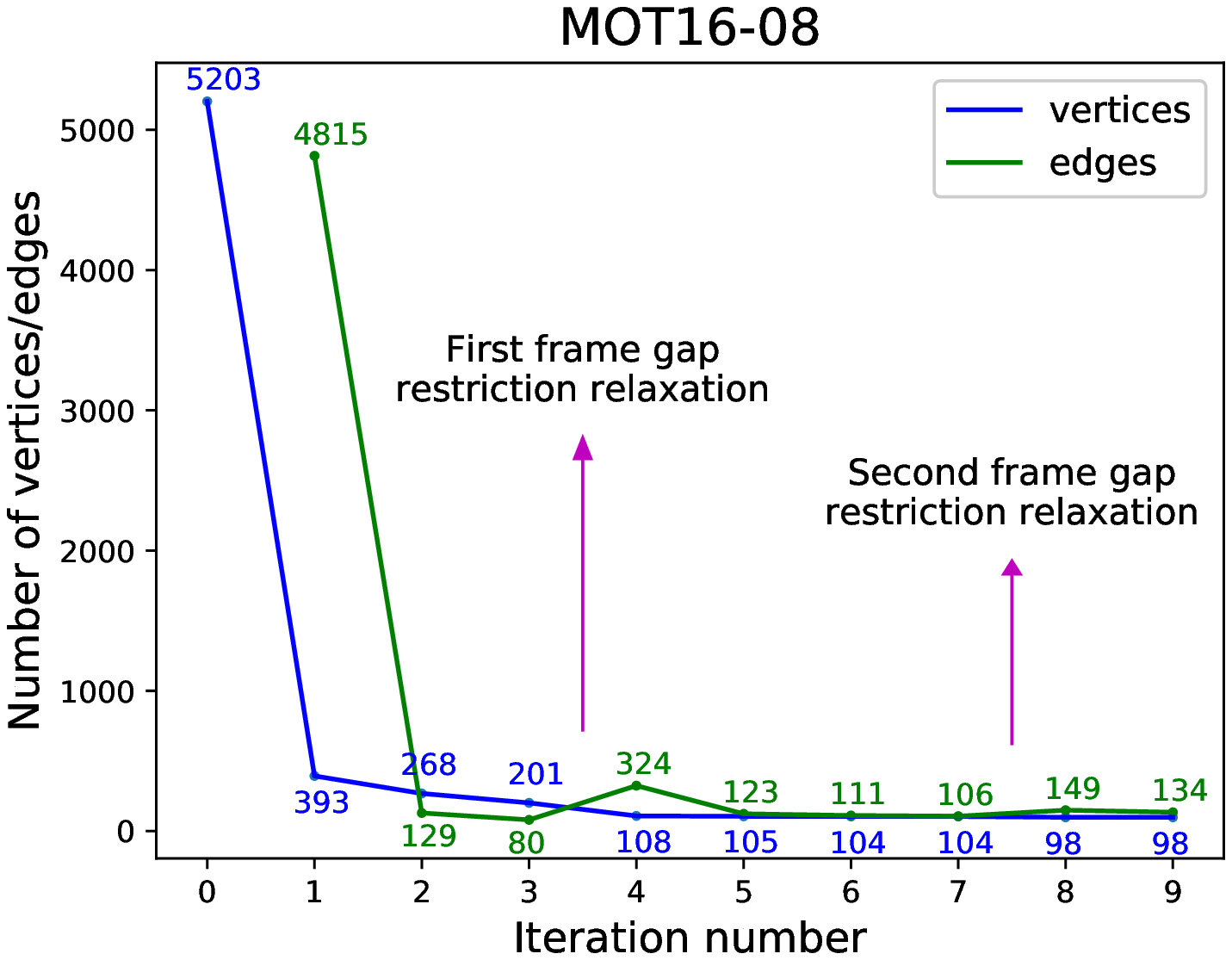}}
  \subfloat{\includegraphics[width=0.35\textwidth]{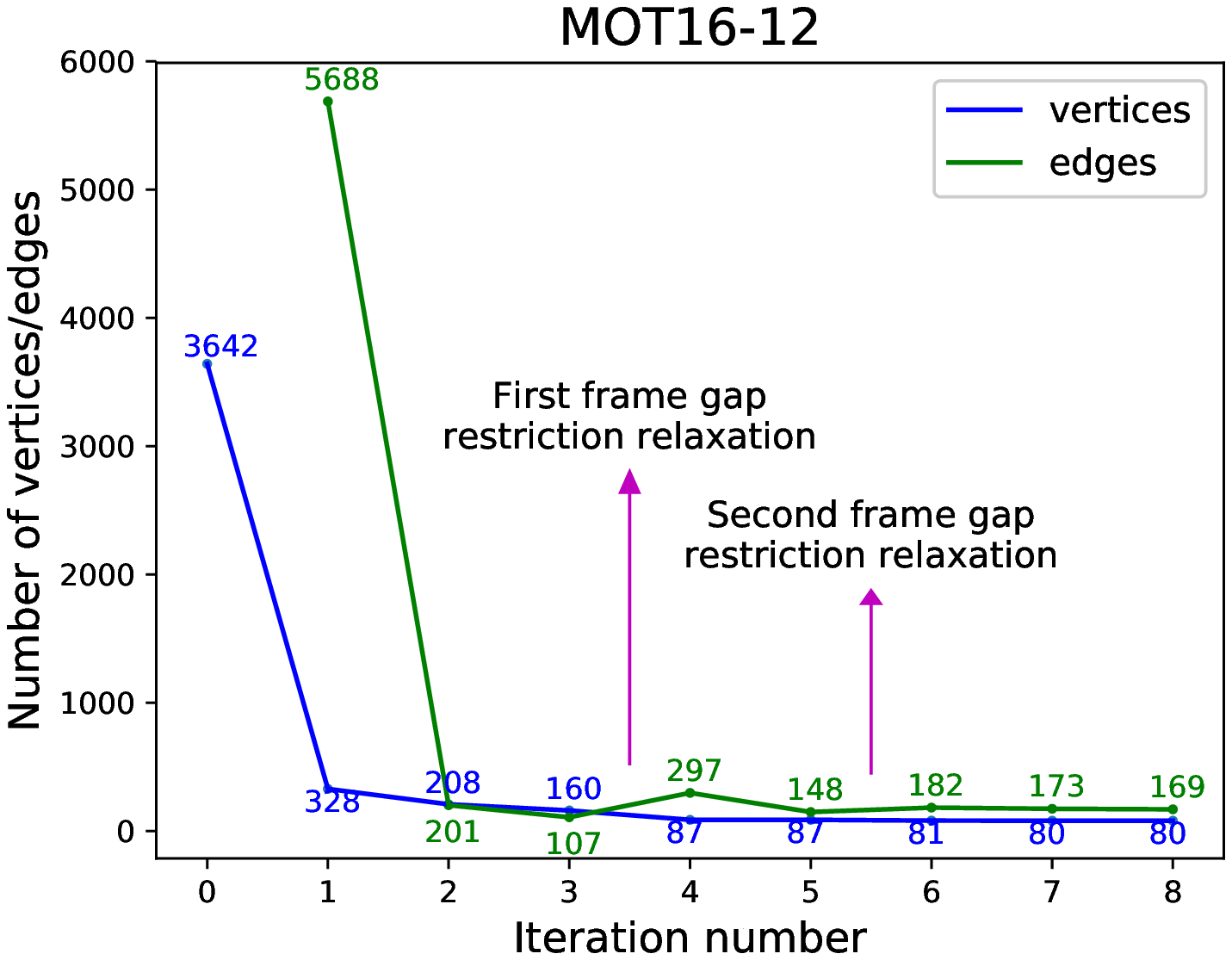}}
  \newline
  \subfloat{\hspace{25pt}\includegraphics[width=0.35\textwidth]{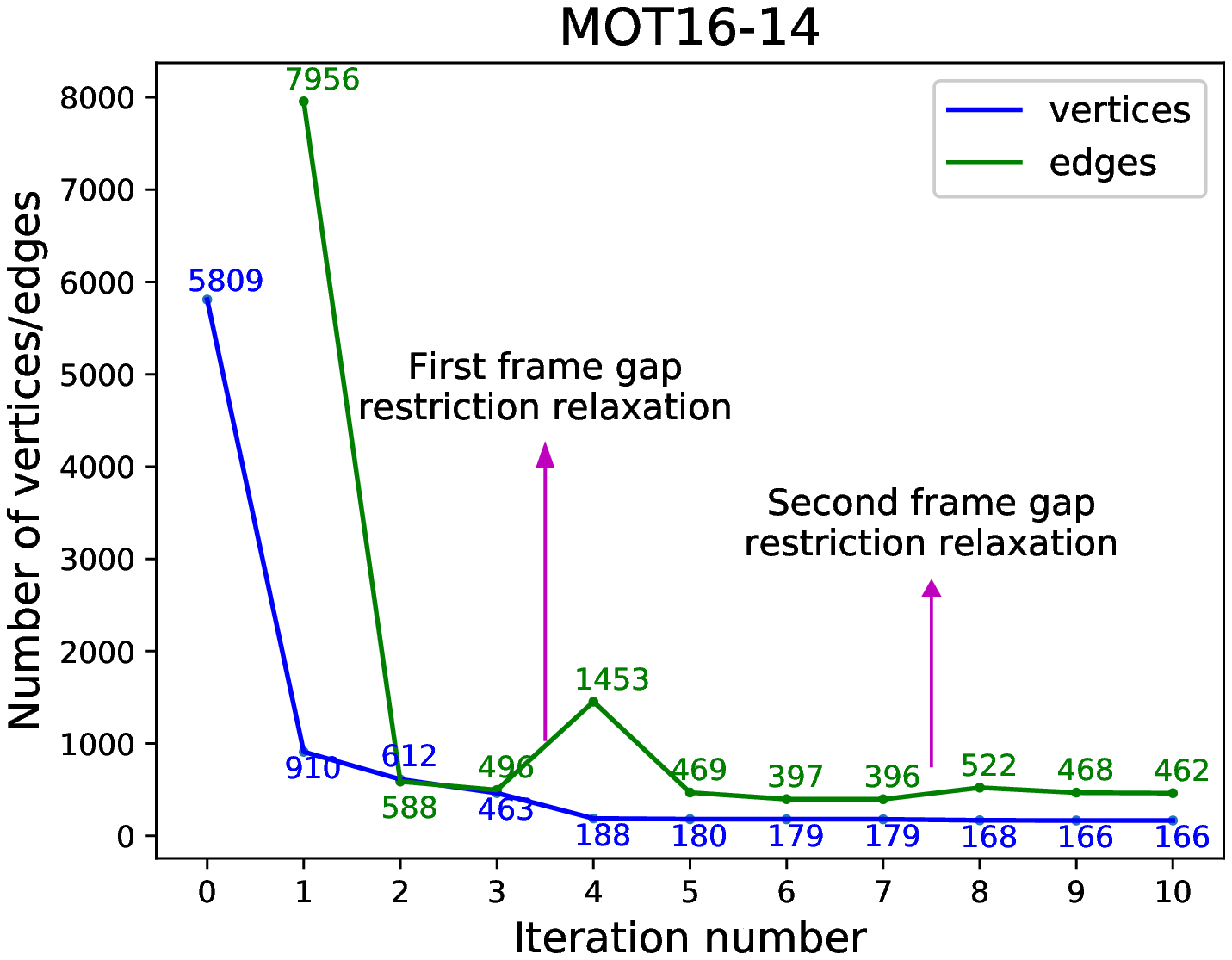}}

\caption{Number of vertices and edges in the graph at each clustering iteration for all MOT16 test datasets.}
\vspace{10pt}
\label{fig:hc_stats}
\end{figure*}

In Fig. \ref{fig:hc_stats}, the number of vertices and edges in the graph for each iteration of the clustering process, and for each test dataset in the MOT16 challenge are illustrated. The key observations from these results are:

\begin{itemize}
\item The number of vertices is, on average, reduced by approximately 90\% after the first iteration, suggesting that a significant part of the clustering is already complete. This behavior is encouraging because it means that the complexity of the graph is greatly reduced after just one iteration.

\item The number of edges and vertices are roughly of the same order. This shows the effectiveness of the edge creation scheme described earlier.

\item The number of iterations required for convergence under the currently employed clustering scheme is 8-12. This is despite the fact that the test dataset videos were captured in varying environments, and have different numbers of detections and frames, and different detection densities (average number of detections per frame).

\item Recall that the criteria for determining the maximum allowable frame gap between tracklets was relaxed in two steps: the first relaxation occurs in the fourth iteration, when the maximum allowed frame gap is increased from 4, to 4 times the tracklet length. This is reflected by the spike in the number of edges created in the fourth iteration. The second relaxation comes when the algorithm initially converges, after which edges are allowed to span 6 times the tracklet length. Here, it is again observable that the number of edges increases. Moreover, the number of vertices usually decreases after this relaxation, even though the algorithm had converged under the previous criteria. The iteration number of both relaxations is marked in magenta colored arrows on the graphs.
\end{itemize}

\section{Timing Analysis}
\label{app:timing_analysis}

\begin{figure*}
\centering
\includegraphics[width=0.4\linewidth]{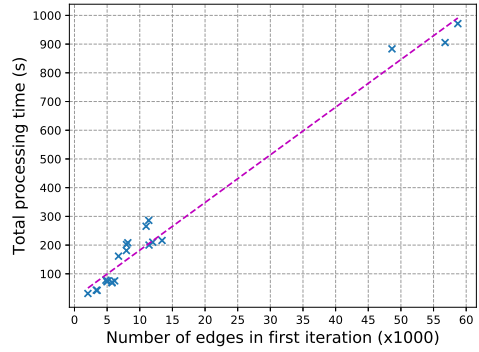}
\caption{Plot of processing time against number of edges in the first iteration for all MOT17 test datasets.}
\label{fig:time_analysis}
\end{figure*}

A common concern with any framework that employs graph optimization is its scalability. Fortunately, we observed that the total computational time (including inference and clustering) required to process each dataset is strongly correlated with the number of edges created in the first iteration. In Fig. \ref{fig:time_analysis}, a scatter plot is drawn of the processing time required for each MOT17 test dataset against the number of edges created in the first iteration. The line of best fit between all the points is also plotted in magenta. It can be seen that despite the varying nature of the datasets, there is a strong linear relation between the two parameters. This trend offers strong empirical support for the scalability of our framework.

All results were obtained on a desktop system with an Intel Xeon E5-1620 CPU running at 3.5GHz with 16GB RAM, and an Nvidia GTX TITAN X GPU.

\section{MOT Challenge Results}
\label{app:mot16_results}

\begin{figure*}
\captionsetup[subfloat]{farskip=1pt,captionskip=1pt}
\centering
  \subfloat{\includegraphics[width=\imgwidth\textwidth]{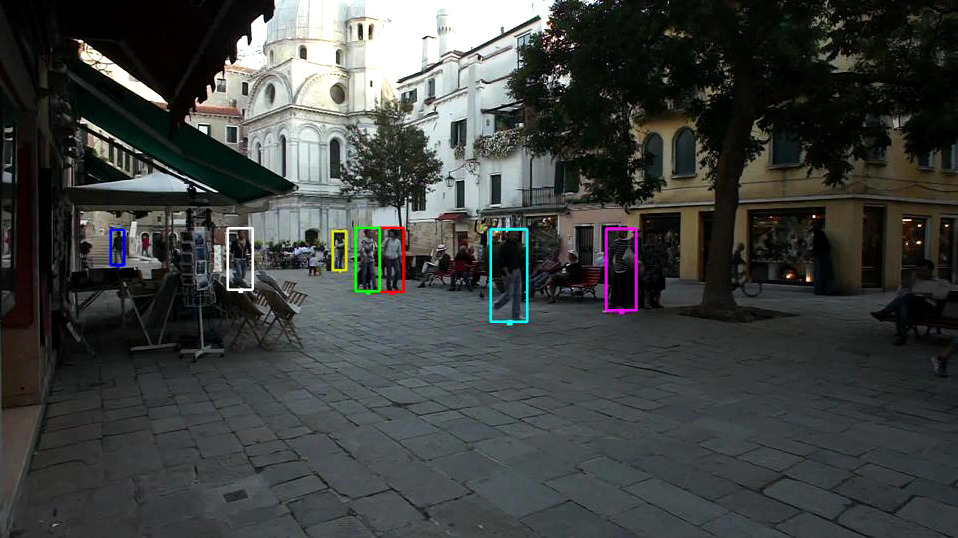}}
  \hspace{5pt}
  \subfloat{\includegraphics[width=\imgwidth\textwidth]{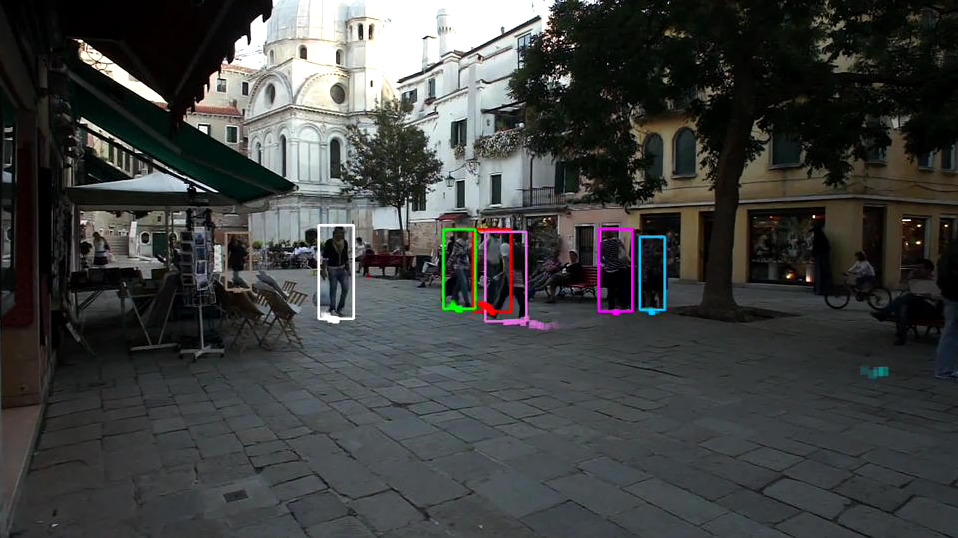}}
  \hspace{5pt}
  \subfloat{\includegraphics[width=\imgwidth\textwidth]{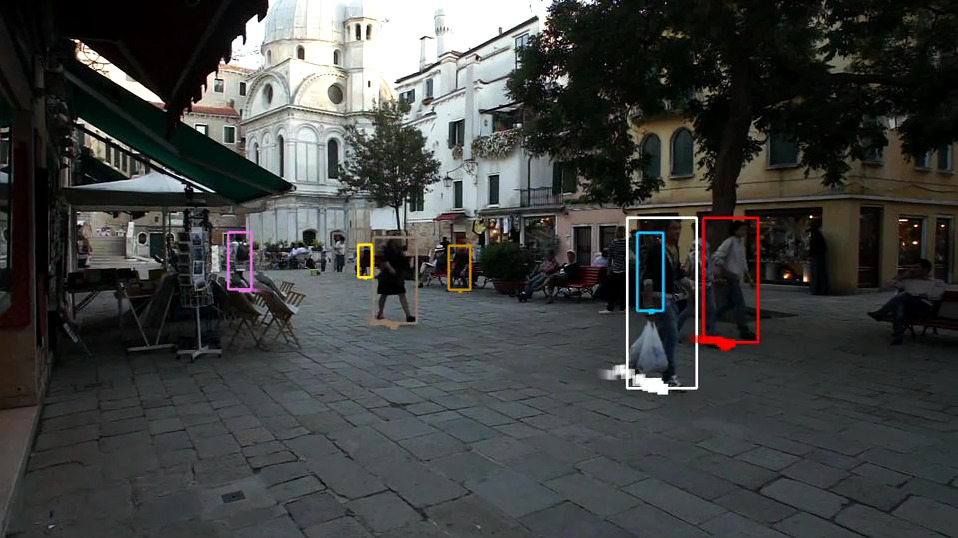}}
  \hspace{5pt}
  \newline
  \subfloat{\includegraphics[width=\imgwidth\textwidth]{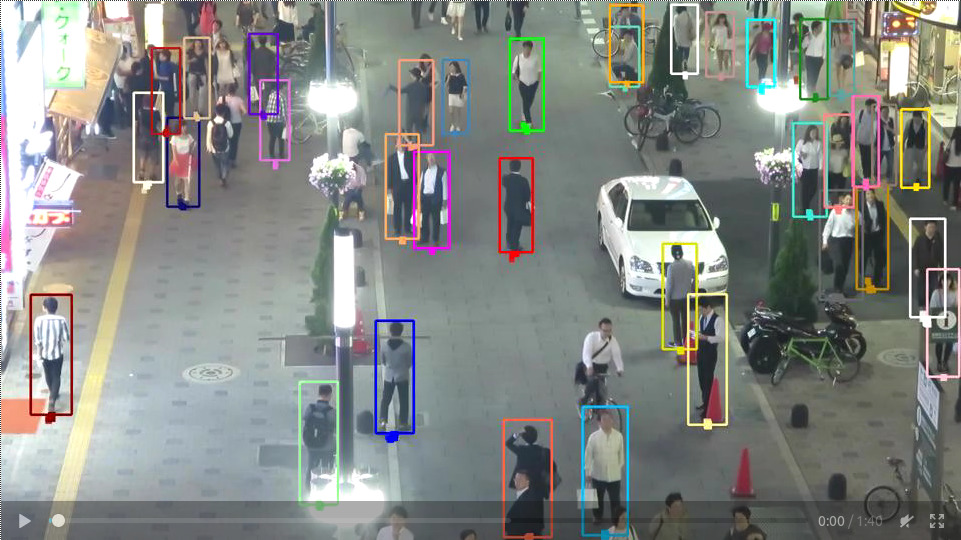}}
  \hspace{5pt}
  \subfloat{\includegraphics[width=\imgwidth\textwidth]{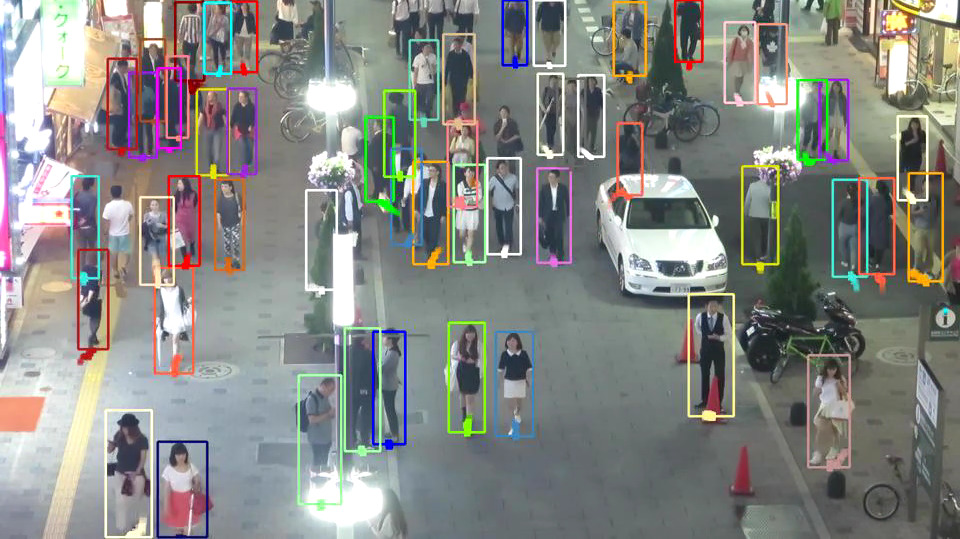}}
  \hspace{5pt}
  \subfloat{\includegraphics[width=\imgwidth\textwidth]{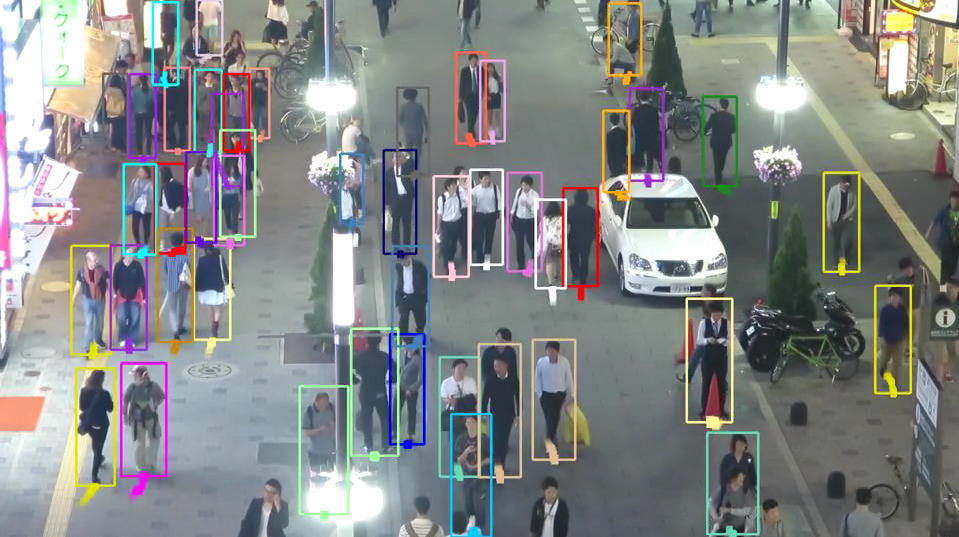}}
  \hspace{5pt}
  \newline
  \subfloat{\includegraphics[width=\imgwidth\textwidth]{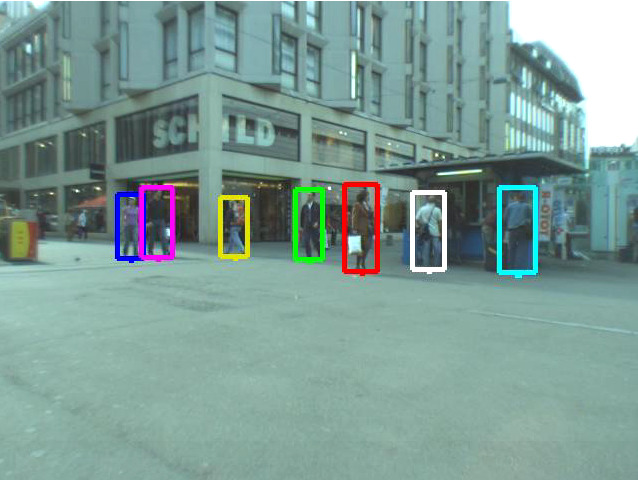}}
  \hspace{5pt}
  \subfloat{\includegraphics[width=\imgwidth\textwidth]{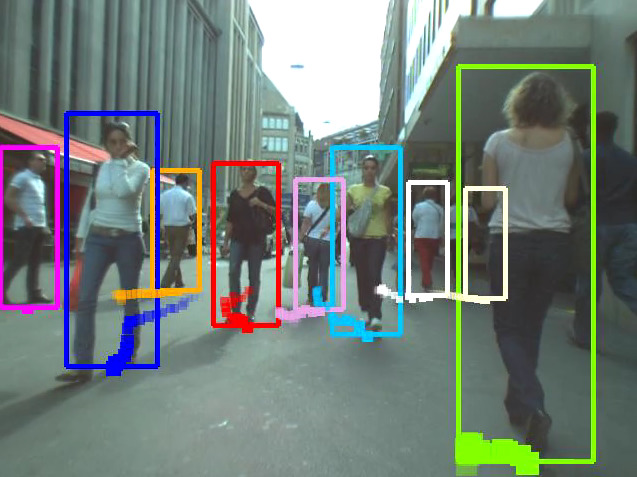}}
  \hspace{5pt}
  \subfloat{\includegraphics[width=\imgwidth\textwidth]{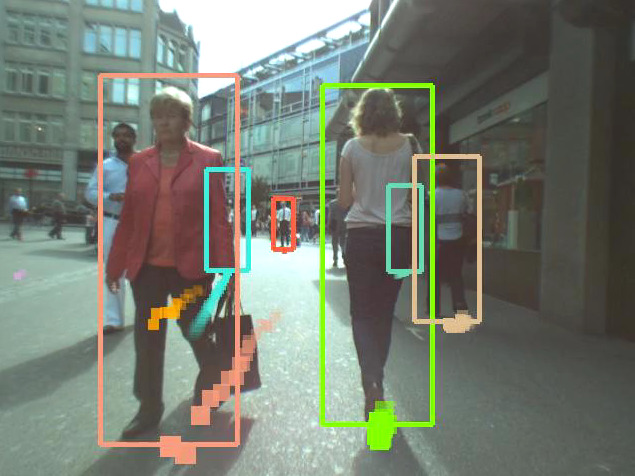}}
  \hspace{5pt}
  \newline
  \subfloat{\includegraphics[width=\imgwidth\textwidth]{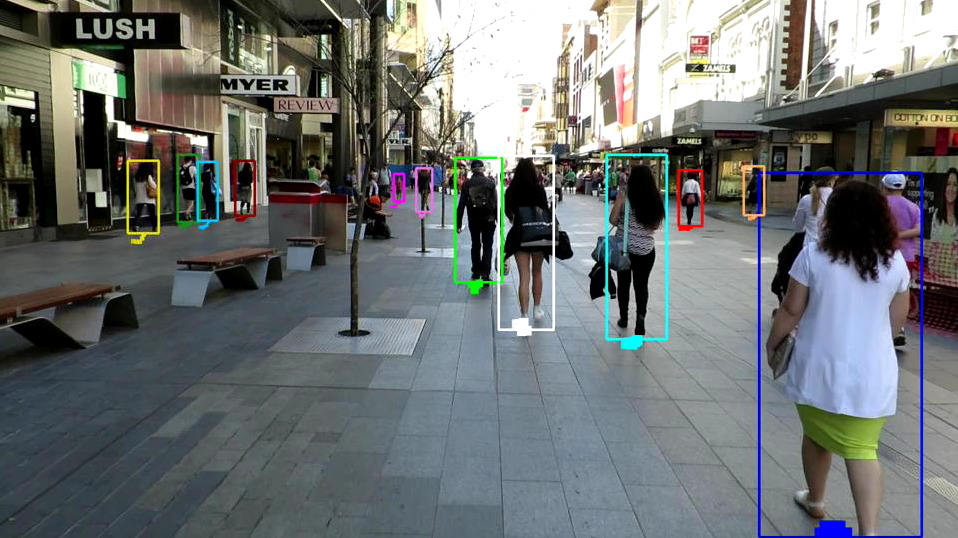}}
  \hspace{5pt}
  \subfloat{\includegraphics[width=\imgwidth\textwidth]{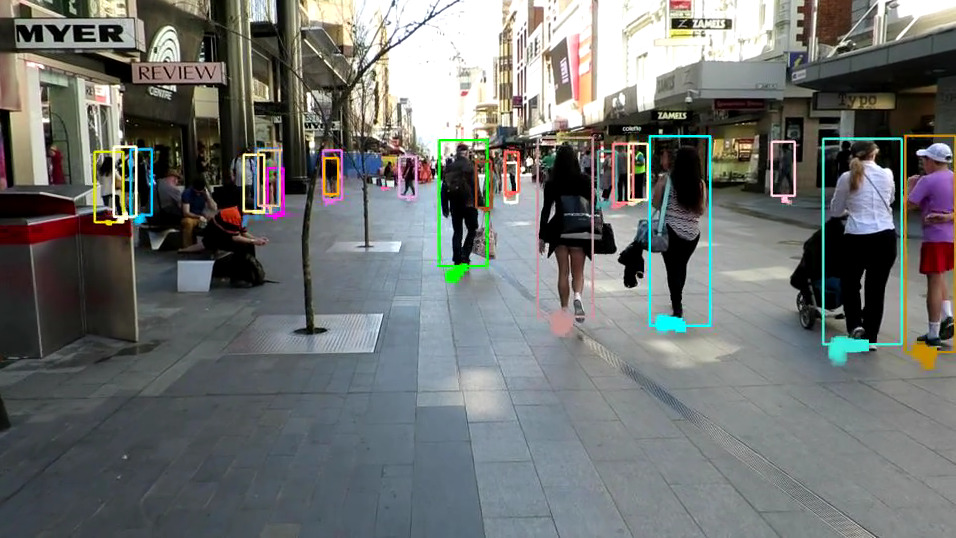}}
  \hspace{5pt}
  \subfloat{\includegraphics[width=\imgwidth\textwidth]{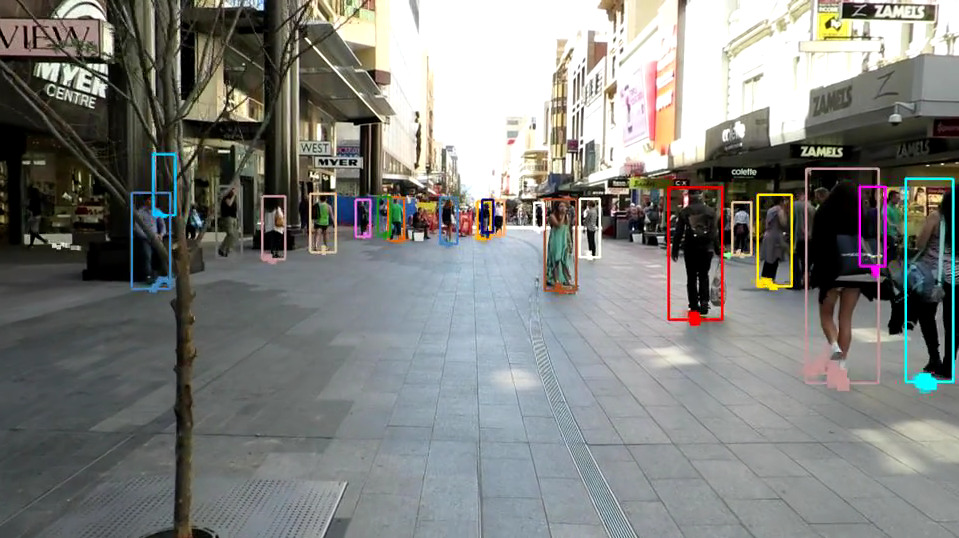}}
  \hspace{5pt}
  \newline
  \subfloat{\includegraphics[width=\imgwidth\textwidth]{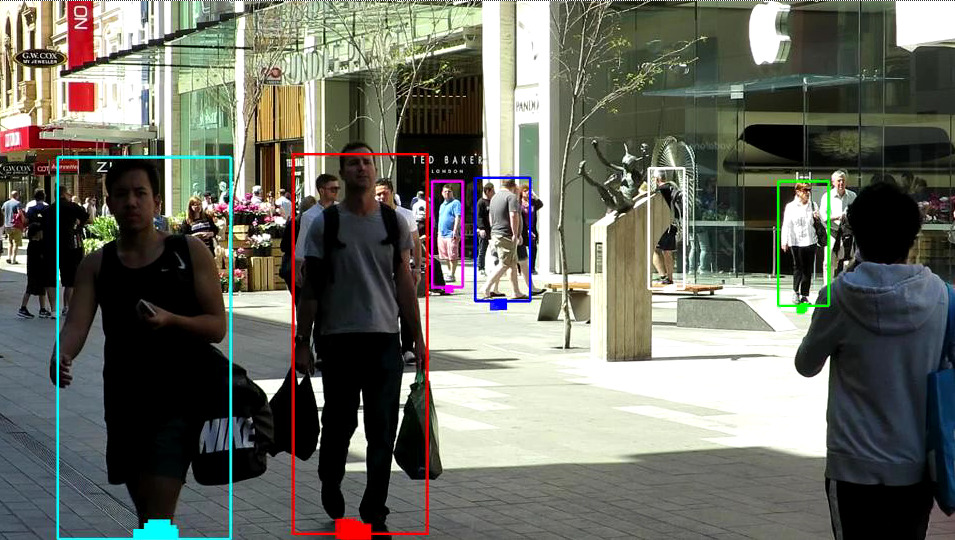}}
  \hspace{5pt}
  \subfloat{\includegraphics[width=\imgwidth\textwidth]{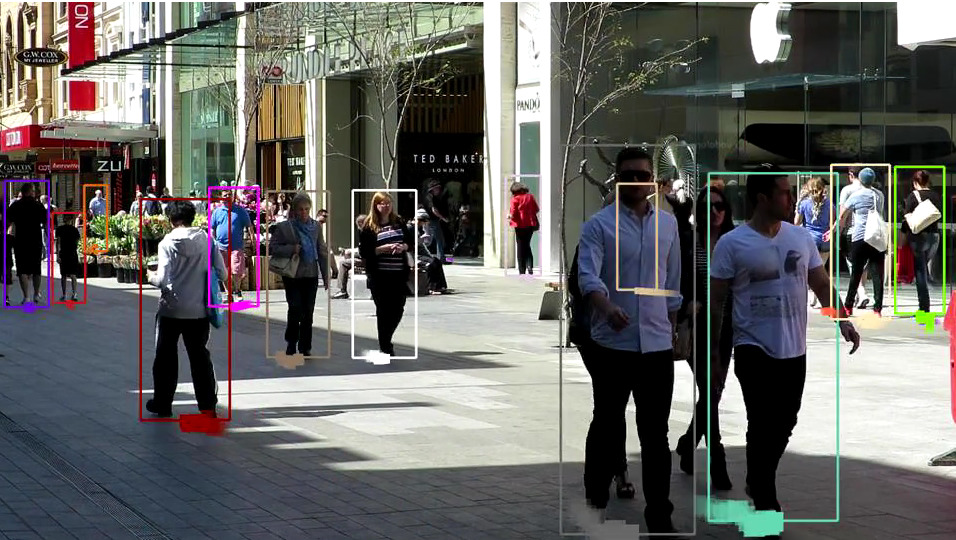}}
  \hspace{5pt}
  \subfloat{\includegraphics[width=\imgwidth\textwidth]{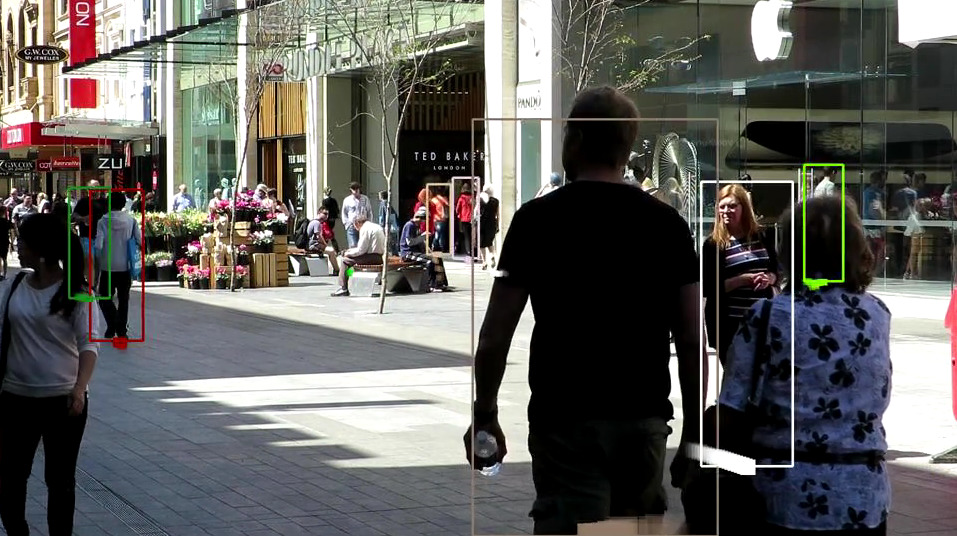}}
  \hspace{5pt}
  \newline
  \subfloat{\includegraphics[width=\imgwidth\textwidth]{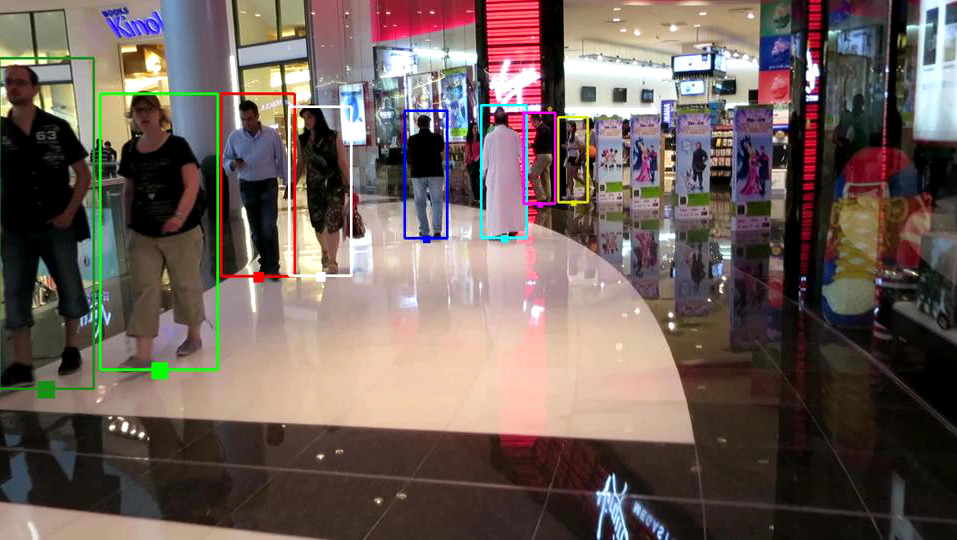}}
  \hspace{5pt}
  \subfloat{\includegraphics[width=\imgwidth\textwidth]{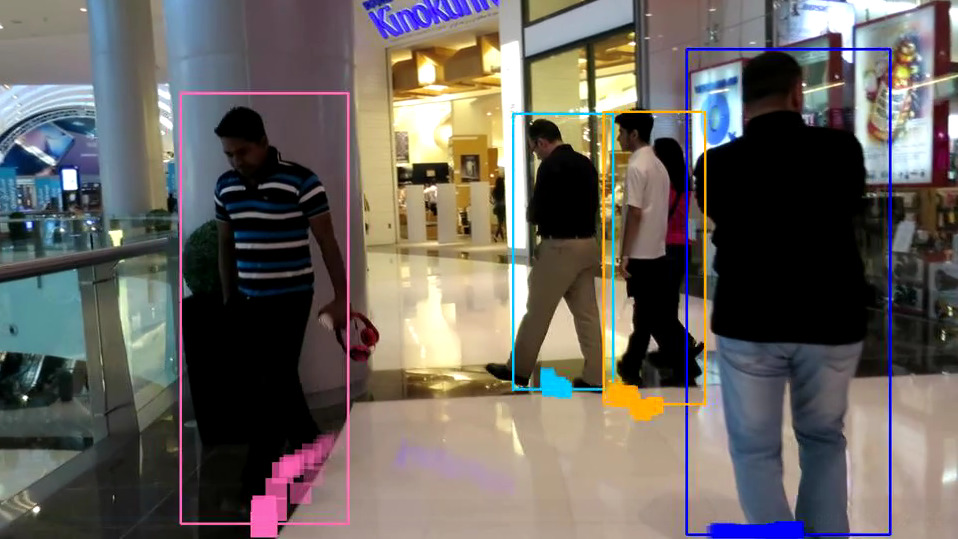}}
  \hspace{5pt}
  \subfloat{\includegraphics[width=\imgwidth\textwidth]{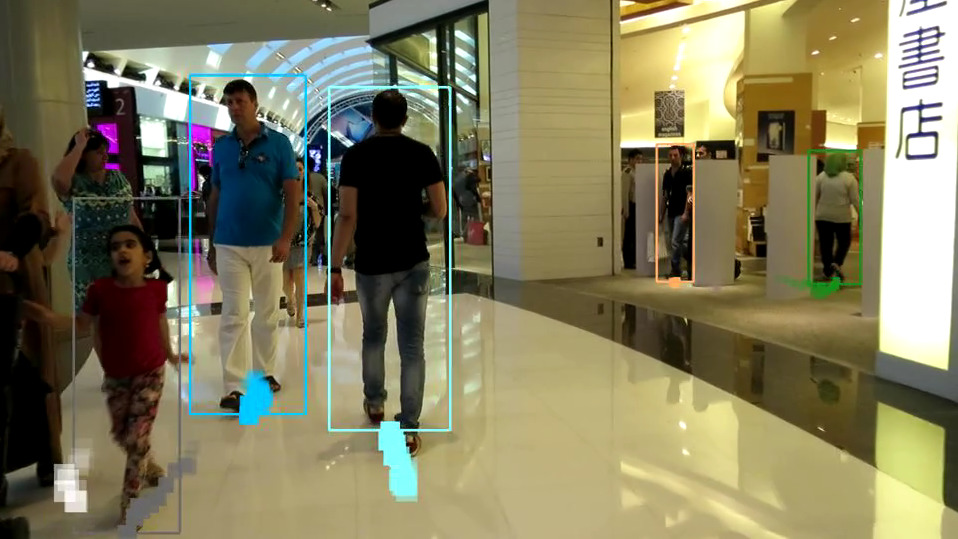}}
  \hspace{5pt}
  \newline
  \subfloat{\includegraphics[width=\imgwidth\textwidth]{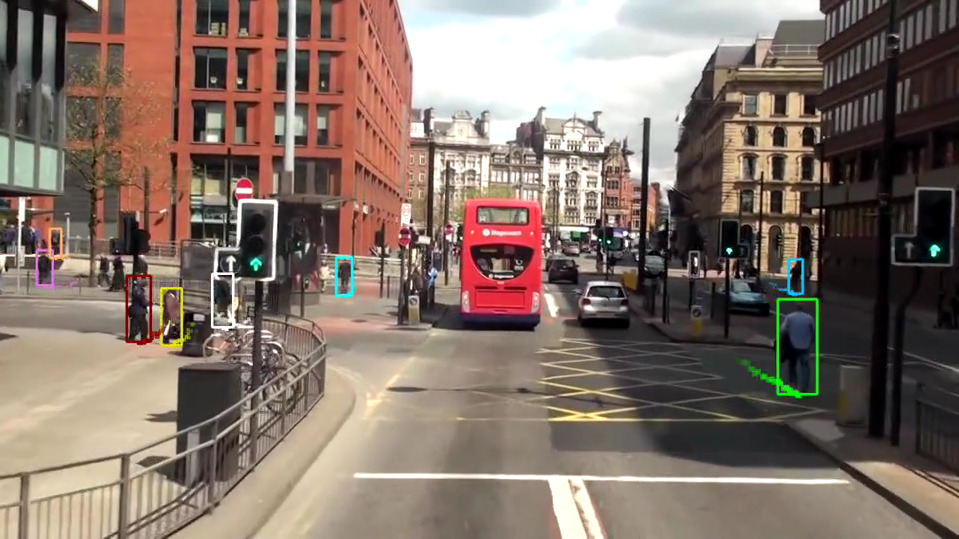}}
  \hspace{5pt}
  \subfloat{\includegraphics[width=\imgwidth\textwidth]{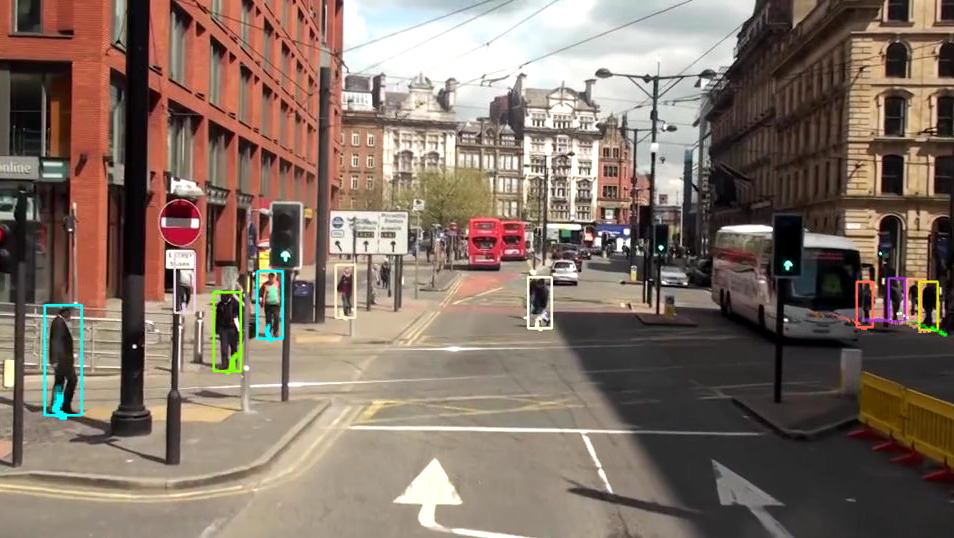}}
  \hspace{5pt}
  \subfloat{\includegraphics[width=\imgwidth\textwidth]{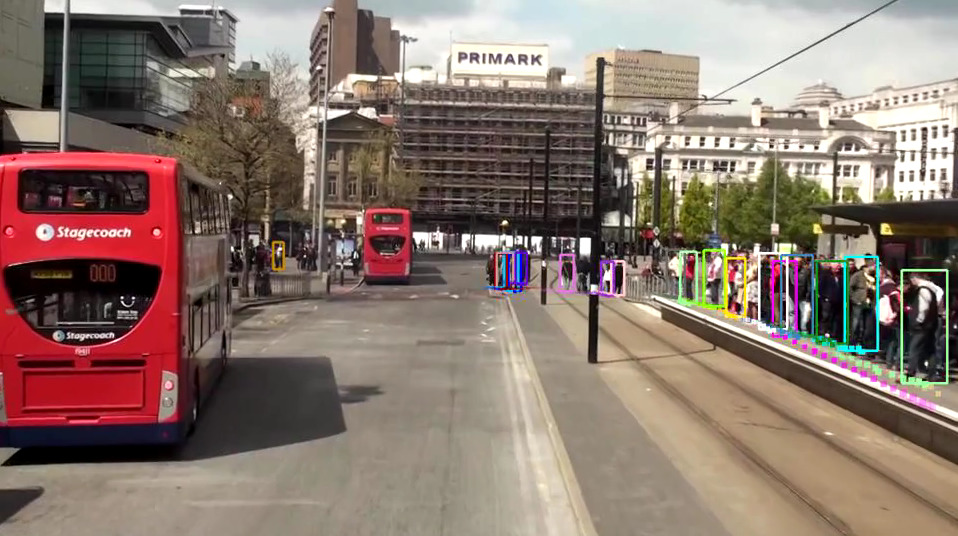}}
  \hspace{5pt}
  \newline

\caption{Screenshots of annotated video sequences from the MOT16 test datasets. Different colors correspond to different person IDs.}
\label{fig:annotated_videos}
\end{figure*}

The performance metrics for each video sequence in the MOT16~\cite{milan2016mot16} test dataset are provided in Table \ref{tb:mot16_detailed}. In Fig. \ref{fig:annotated_videos}, screenshots of annotated video sequences of the MOT16 test dataset are given. Three screenshots from each of the seven videos are given in each row of the tiled figure. The results provided in the two tables, as well as the full video sequences are available online for both MOT16\footnote{\url{https://motchallenge.net/tracker/HDTR_16}} and MOT17 challenges\footnote{\url{https://motchallenge.net/tracker/HDTR_17}}.

\begin{figure}[H]
\centering
\includegraphics[width=\linewidth]{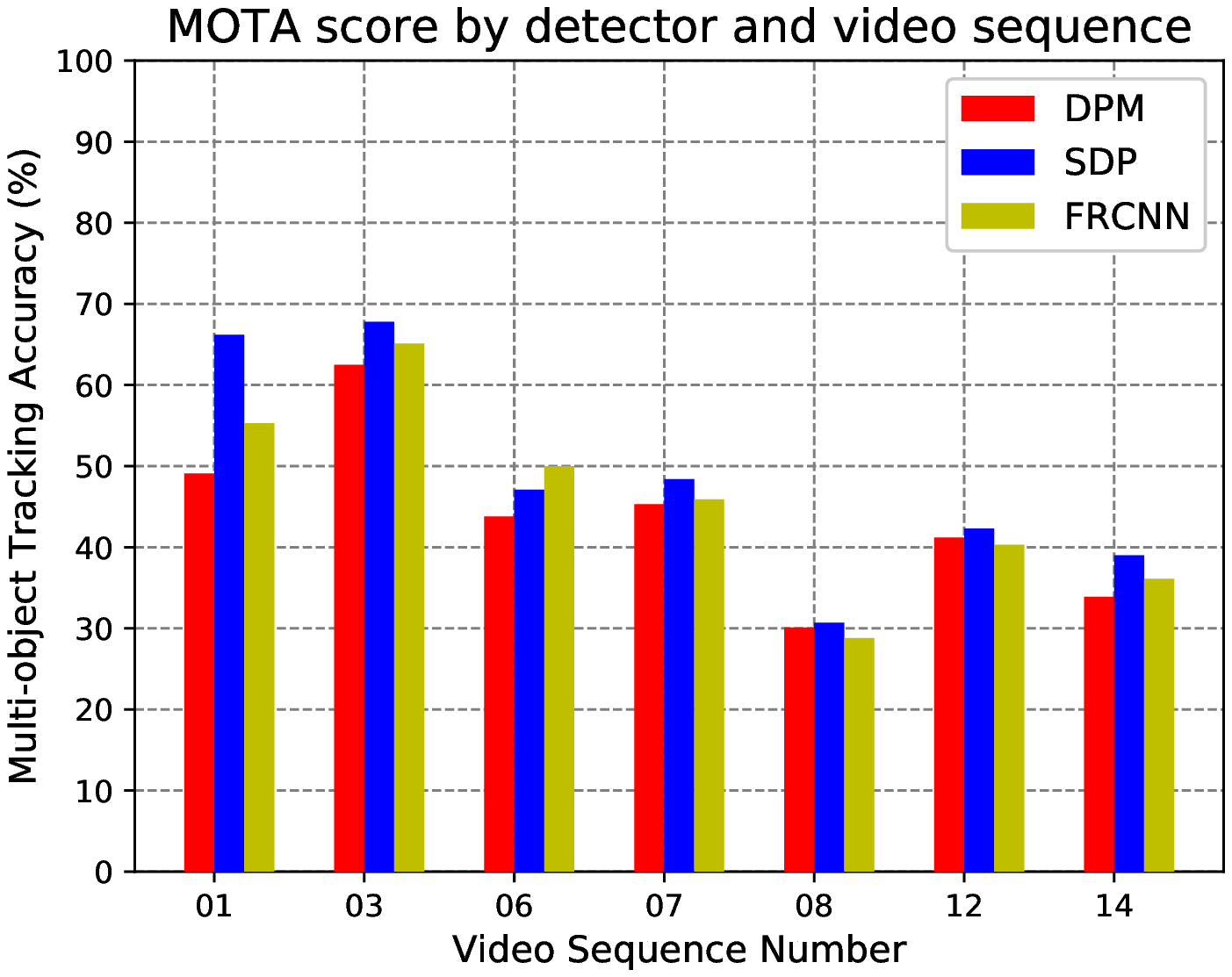}
\caption{Example of training sample creation for RNN.}
\label{fig:mot17_bar}
\end{figure}

As mentioned in the main text, the MOT17 challenge contains the same video sequences as MOT16, but with a different ground truth, and with three sets of detections which are produced by different publicly available object detectors: DPM~\cite{dpm_detector}, SDP~\cite{sdp_detector} and FRCNN~\cite{frcnn_detector}. In Fig.~\ref{fig:mot17_bar}, a bar plot shows how the MOT Accuracy score (as defined by MOT Clear metrics~\cite{motclear2008}) varies for each of the seven video sequences depending on which detector the detections came from.

\begin{table*}
\centering
\small
\begin{tabular}{lccccccccccc}
\hline
Tracker & MOTA$(\%)\uparrow$  & IDF1$(\%)\uparrow$ & MOTP$(\%)\uparrow$   & FAF$\downarrow$  & MT$(\%)\uparrow$  & ML$(\%)\downarrow$ & FP$\downarrow$  & FN$\downarrow$  & ID Sw.$\downarrow$   & Frag$\downarrow$ \\[1ex]
\hline 
MOT16-01 &	49.5 &	38.2 &	80.6 &	0.1 &	30.4 & 	30.4 & 	40 &	3178 &	10 &	15  \\
MOT16-03 &	62.9 &	52.1 &	81.0 &	0.8 &	31.1 & 	16.9 & 	1175 &	37473 &	185 &	262\\
MOT16-06 &	43.2 &	35.2 &	81.2 &	1.3 &	27.1 & 	42.1 & 	1565 &	4898 &	88 &	130\\
MOT16-07 &	46.9 &	42.6 &	79.5 &	1.6 &	20.4 & 	24.1 & 	788 &	7778 &	105 &	143\\
MOT16-08 &	37.7 &	37.2 &	82.1 &	0.3 &	17.5 & 	38.1 & 	163 &	10219 &	45 &	73\\
MOT16-12 &	42.4 &	49.9 &	80.4 &	0.3 &	15.1 & 	48.8 & 	247 &	4482 &	53 &	46\\
MOT16-14 &	34.0 &	33.2 &	78.2 &	1.0 &	7.9 & 	47.0 & 	736 &	11325 &	132 &	164\\
\hline		
\end{tabular}
\caption{Tracking results for each video sequence in the MOT16 test dataset with public detections. $\uparrow$ and $\downarrow$ represent higher is better and lower is better, respectively. The values in bold and blue represent the best and second best performances, respectively.} 
\label{tb:mot16_detailed}
%\end{adjustbox}
\end{table*}

%\end{appendices}

\end{document}